\documentclass[11pt]{article}
\usepackage[left=1.2in,top=1in,bottom=1in,right=1.2in]{geometry}
\usepackage[T1]{fontenc}
\usepackage{amsmath,amssymb,amsfonts,booktabs,caption}
\usepackage{algorithm,algpseudocode,graphicx,multirow,array,makecell,tabularx,adjustbox,float,placeins}
\usepackage[normalem]{ulem}
\usepackage[table]{xcolor}
\usepackage{tikz,lmodern}
\usepackage{microtype}
\usepackage[numbers,sort&compress]{natbib}
\usepackage[colorlinks=true,allcolors=blue]{hyperref}
\newcommand{\exphead}[1]{\noindent\textbf{#1.}}
\newcommand{\varname}[1]{\texttt{\detokenize{#1}}}
\providecommand{\Description}[1]{}
\title{Basin2Vec: Learning Basin-Centric Embeddings for Scalable Hydrological Modeling}
\author{\normalsize Muhammad Talha$^1$ \quad Xiaofeng Liu$^2$ \quad Yuan Yang$^3$\\
\normalsize Pang-Ning Tan$^1$ \quad A. Pouyan Nejadhashemi$^1$\\[8pt]
\normalsize $^1$Michigan State University\\
\normalsize $^2$University of Michigan \quad $^3$University of California, San Diego\\[5pt]
\small\texttt{talhamuh@msu.edu}, \texttt{xiaofliu@umich.edu}, \texttt{yuy068@ucsd.edu}\\
\small\texttt{ptan@msu.edu}, \texttt{pouyan@msu.edu}}
\date{}
\begin{document}
\maketitle
\begin{center}\small Accepted at the 2027 IEEE/CVF Winter Conference on Applications of Computer Vision (WACV).\end{center}

\begin{abstract}
Reliable streamflow prediction across time and in ungauged basins requires representations that capture hydrologically relevant basin characteristics. Existing methods rely on either fixed handcrafted attributes or general-purpose Earth observation embeddings not tailored to basin structure. We introduce \textbf{Basin2Vec}, a self-supervised framework that learns compact basin representations from hydroclimatic and physiographic data without observed streamflow. Basin2Vec organizes representation learning around basin scale and relationships across space and time. Across multiple forecasting architectures, it consistently outperforms predefined attributes and general-purpose geospatial embeddings in temporal generalization and zero-shot spatial transfer. Basin2Vec provides average relative gains of up to 8.3\% temporally and achieves 3.5$\times$ the median predictive performance of the strongest baselines on unseen basins. These results demonstrate the effectiveness of basin-centric pretraining for transferable hydrologic prediction. Code is available at  \url{https://github.com/Muhammad-Talha-MT/basin2vec}.
\end{abstract}

\section{Introduction}

Streamflow, or river discharge, is the volume of water passing through a river cross-section per unit time. Accurate streamflow prediction is critical for flood forecasting, drought monitoring, and water resources planning. However, streamflow observations are available at only a limited and unevenly distributed network of gauging stations \citep{Brunner2021FloodDrought,Samaniego2019HydrologicalForecasts,USGS2024Streamgages}. Therefore, hydrological models must generalize in two ways:
\textit{temporal generalization}, in which they predict future streamflow in
basins seen during training, and \textit{spatial generalization}, in which they predict streamflow in basins that were not used for training
~\citep{Kratzert2019PUB,Fang2022DataSynergy}. Accurate prediction in both settings requires a robust basin representation. In particular, the representation must capture the \textit{dynamic meteorological inputs}, such as precipitation and temperature, along with a \textit{static basin descriptor} that distinguishes characteristics of the target basin from others. Such a descriptor must summarize the factors governing how a basin converts weather into streamflow, including climate, terrain, soils, land cover, subsurface properties, and drainage area \citep{Addor2017CAMELS,Falcone2017GAGES,McMillan2021HydrologicSignatures,Tarasova2024Catchment}. An informative basin representation is particularly important for spatial generalization because it enables the model to identify similarities and differences between observed and previously unseen basins.

Regional streamflow models commonly represent each basin as a fixed-length feature vector derived from datasets such as CAMELS and GAGES-II~\citep{Addor2017CAMELS,Falcone2010GAGES,Falcone2017GAGES,Kratzert2019EntityAwareLSTM}. While these compact attributes may summarize the spatial characteristics of the basin using statistics such as mean elevation or forest fraction, they do not capture the spatial variation of these characteristics within the basin, causing the downstream models to be oblivious to their spatial heterogeneity. Gridded datasets such as CAMELS-SPAT preserve this information but introduce high-dimensional spatial inputs~\citep{Knoben2025CAMELSSPAT}. This motivates learning compact representations directly from basin-aligned geospatial data.

% Score marker: 3 = High, 2 = Moderate, 1 = Low
\newcommand{\score}[1]{%
\begin{tikzpicture}[baseline=-0.55ex]
\foreach \i in {1,...,3}{
\pgfmathtruncatemacro{\isgreen}{\i <= #1 ? 1 : 0}
\ifnum\isgreen=1
\fill[teal!70!black] ({(\i-1)*0.16},0) circle (0.07);
\else
\fill[red!75!black] ({(\i-1)*0.16},0) circle (0.07);
\fi
}
\end{tikzpicture}%
}

\begin{table}[!htbp]
\centering
\scriptsize
\setlength{\tabcolsep}{3pt}
\renewcommand{\arraystretch}{1.10}
\setlength{\abovecaptionskip}{2pt}
\setlength{\belowcaptionskip}{3pt}
\caption{
Comparison of basin representations across three design requirements.
Scores indicate High, Moderate, or Low support.
}
\label{tab:representation_motivation}
\begin{tabularx}{\columnwidth}{@{}
>{\raggedright\arraybackslash}X
>{\centering\arraybackslash}p{0.15\columnwidth}
>{\centering\arraybackslash}p{0.20\columnwidth}
>{\centering\arraybackslash}p{0.25\columnwidth}
@{}}
\toprule
\textbf{Approach}
& \makecell{\textbf{Basin-}\\\textbf{indexed}}
& \makecell{\textbf{Spatial \&}\\\textbf{scale-aware}}
& \makecell{\textbf{Hydrologic}\\\textbf{relation-aware}} \\
\midrule

CAMELS Attr.~\citep{Addor2017CAMELS}
& \score{4} & \score{1} & \score{2} \\

AlphaEarth~\citep{Brown2025AlphaEarth}
& \score{1} & \score{2} & \score{0} \\

TESSERA~\citep{Feng2025TESSERA}
& \score{1} & \score{2} & \score{0} \\

SatCLIP~\citep{Klemmer2025SatCLIP}
& \score{1} & \score{2} & \score{0} \\

\midrule
\rowcolor{gray!15}
\textbf{Basin2Vec (Ours)}
& \score{3} & \score{3} & \score{3} \\

\bottomrule
\end{tabularx}

\vspace{3pt}
\begin{adjustbox}{max width=\linewidth}
\begin{tabular}{@{}ccc@{}}
\score{3}~High
& \hspace{8pt}\score{2}~Moderate
& \hspace{8pt}\score{1}~Low
\end{tabular}
\end{adjustbox}

\end{table}

General-purpose Earth observation (EO) embeddings provide reusable representations of satellite observations and geographic locations~\citep{Brown2025AlphaEarth,Feng2025TESSERA,Klemmer2025SatCLIP}. However, they are pretrained on pixels, image patches, or coordinates rather than drainage basins, requiring post hoc aggregation to construct basin descriptors. Moreover, their pretraining objectives do not explicitly encode physical basin scale or hydrologically meaningful relationships among basins.

Together, these limitations motivate three design requirements for basin-centric representation, as summarized in Table~\ref{tab:representation_motivation}. First, the representation should be \textit{basin-indexed}, meaning that its entities correspond to drainage basins rather than individual pixels or coordinates. Second, the learned embeddings must be \textit{spatial- and scale-aware}, meaning that it learns from within-basin spatial inputs while retaining information about physical basin scale. Finally, it should be \textit{hydrologic relation-aware}, meaning that the learned embedding geometry reflects the hydrological and physiographic relations among the basins.

To meet these requirements, we introduce \textbf{Basin2Vec}. Our contributions
are threefold. \textbf{(1)} We formulate basin-centric self-supervision around a
\textit{basin-year} entity, producing reusable descriptors from gridded
hydroclimatic and physiographic data without observed streamflow.
\textbf{(2)} We propose a \emph{scale-aware encoder} that preserves spatial and
physical-scale information, together with a \emph{relation-aware contrastive
objective} that models temporal, overlapping, and nested basin relationships.
\textbf{(3)} We evaluate the resulting descriptors with five forecasting
architectures under temporal generalization and zero-shot spatial transfer.
Basin2Vec consistently outperforms predefined attributes and general-purpose EO
embeddings, with average relative temporal gains of up to 8.3\% and 3.5$\times$
the median predictive performance of the strongest alternative on unseen basins.

\section{Related Work}

\paragraph{Basin characterization in regional hydrological models.}
Large-sample hydrological models are commonly conditioned on curated basin descriptors from CAMELS or GAGES-II~\citep{Newman2015CAMELS,Addor2017CAMELS,Falcone2010GAGES,Falcone2017GAGES}. This paradigm, including entity-aware and regional LSTMs, has demonstrated the value of static attributes for shared prediction across multiple basins~\citep{Kratzert2019PUB,Kratzert2019EntityAwareLSTM,Fang2022DataSynergy}. CAMELS-SPAT extends this foundation with basin-aligned rasters of climatic and physiographic variables~\citep{Knoben2025CAMELSSPAT}, but using them directly requires each downstream model to process high-dimensional spatial inputs. 

\paragraph{Earth-observation representation learning.}
General-purpose EO representation-learning methods derive transferable features from remotely sensed or geographic data. SatMAE learns visual representations from satellite image patches~\citep{Cong2022SatMAE}; AlphaEarth and TESSERA learn geospatial representations from EO observations; and SatCLIP learns location representations from geographic coordinates~\citep{Brown2025AlphaEarth,Feng2025TESSERA,Klemmer2025SatCLIP}. Recent studies have used these representations as inputs to hydrological models~\citep{Ou2026SatelliteHydrology,Lin2026SatelliteStreamflow,Qu2026AlphaEarthHydrology}. Because these representations are natively associated with pixels, patches, or coordinates, their application to hydrological basins typically requires post hoc spatial aggregation. 

\paragraph{Hydrology-specific representation learning.}
Previous hydrology-specific learning methods have used encoder-decoder or autoencoder to infer hydrological signatures and landscape features from meteorological and streamflow time series~\citep{Botterill2023HydrologicSignatures,Bassi2024LandscapeFeatures}. KG-SSL similarly learns basin representations from basin-level meteorological inputs and observed streamflow, treating samples from the same basin across different years as related and samples from different basins as unrelated~\citep{ghosh2022robust}. However, these methods rely on observed streamflow during the representation learning, limiting their applicability to ungauged basins.

\section{Basin2Vec}
\label{sec:method}

\subsection{Overview}
\label{sec:basin2vec_overview}
\begin{figure}[!htbp]
\centering
\includegraphics[width=0.98\textwidth, trim={3.21cm 6.30cm 3.1cm 6.4cm}, clip]{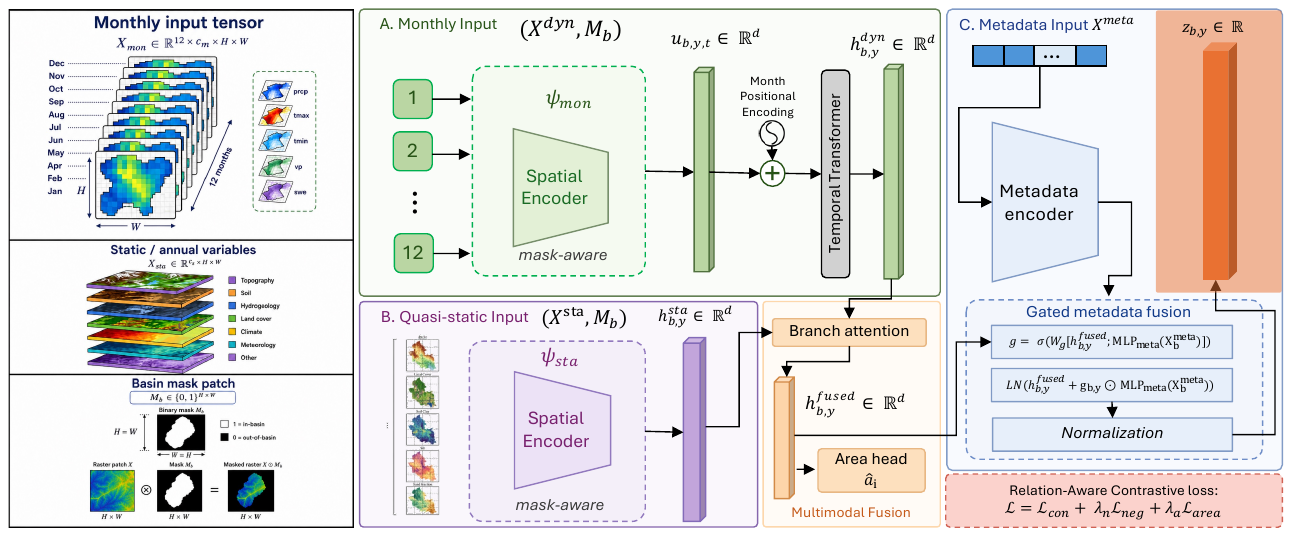}
\captionsetup{skip=2pt}
\caption{
\textbf{Overview of Basin2Vec.} The scale-aware encoder maps monthly and quasi-static basin-aligned grids,
masks, and metadata to normalized basin-year embeddings $z_{b,y}$ through
mask-aware encoding, multimodal fusion, and area regularization.
}
\label{fig:basin2vec_encoder}

\end{figure}

Basin2Vec learns basin-level representations from basin-year entities.
Each entity corresponds to a basin-year tuple, $(b,y)$, along with its monthly hydroclimatic rasters $X_{b,y}^{\mathrm{dyn}}$, static or annual spatial variables $X_{b,y}^{\mathrm{sta}}$,  binary basin mask $M_b$, and basin metadata $X_b^{\textrm{meta}}$. Note that each basin has been resampled to a fixed $H\times W$ patch using the approach described in~\citep{talha2026basingrid}, although its boundary is irregular and its physical area may vary substantially. Details of the resampling approach are given in Appendix~A.2. The monthly input has shape $X_{b,y}^{\mathrm{dyn}}\in\mathbb{R}^{12\times C_d\times H\times W}$, where $C_d$ is the number of hydroclimatic variables. The quasi-static input has shape
$X_{b,y}^{\mathrm{sta}}\in\mathbb{R}^{C_s\times H\times W}$, where $C_s$ is the number of static or annually varying physiographic variables. As shown in Figure~\ref{fig:basin2vec_encoder}, Basin2Vec uses a \textbf{scale-aware encoder} to map these masked basin-aligned inputs to a normalized basin-year embedding $z_{b,y}\in\mathbb{R}^{d}$, and a \textbf{relation-aware contrastive objective} to structure the embedding space using same-basin, temporal, spatial overlap, nesting, and physiographic relations. After pretraining, the basin-year embeddings ${z_{b,y}:y\in\mathcal{Y}_b}$ are averaged across the available years $\mathcal{Y}_b$ to obtain the fixed basin-level descriptor $\bar{z}_b\in\mathbb{R}^{d}$ for downstream streamflow prediction. 

\subsection{Scale-Aware Basin Encoder}
\label{sec:basin_encoder}

The encoder generates the basin-year embedding $z_{b,y}$ by processing the inputs defined in Section~\ref{sec:basin2vec_overview} according to the following steps: mask-aware spatial encoding of the monthly and quasi-static inputs, fusion of the two resulting representations, and incorporation of basin metadata. Since the basins have been resampled to the same spatial dimensions, which removes the information about its physical size, an auxiliary drainage-area objective is also applied during pretraining.

\exphead{Mask-Aware Spatial Encoding} Each basin-aligned input contains grid cells both inside and outside the
irregular basin boundary. To prevent out-of-basin cells from affecting the
representation, Basin2Vec applies the binary mask $M_b$ during spatial encoding
and pooling. Given a multi-channel input tensor $X$, whose last two dimensions
correspond to the $H\times W$ spatial grid, a convolutional neural network (CNN) is used to extract a spatial feature map $F$ from the masked input. The feature map is then aggregated
via masked average pooling $\mathrm{AvgPool}_{M_b}$ and mask-normalized
attention pooling $\mathrm{AttnPool}_{M_b}$, as shown below:

\begin{equation}
\begin{aligned}
u
&=
\psi(X,M_b)
\\
&\equiv
\mathrm{MLP}
\left(
\left[
\mathrm{AvgPool}_{M_b}(F);
\mathrm{AttnPool}_{M_b}(F)
\right]
\right),
\end{aligned}
\label{eq:masked_spatial_encoder}
\end{equation}
where $\psi$ denotes the mask-aware spatial encoder, $F = \mathrm{CNN}\left(X\odot M_b\right)$ denotes the masked spatial feature map, and $\odot$ denotes element-wise multiplication.
$\mathrm{AvgPool}_{M_b}$ averages the features across all valid basin cells,
whereas $\mathrm{AttnPool}_{M_b}$ assigns learned weights to these cells and
normalizes the weights only within the basin boundary. The two pooled summaries
are concatenated and transformed by an MLP to produce the spatial basin token
$u\in\mathbb{R}^{d}$.

The mask-aware spatial encoder $\psi$ is applied separately to the dynamic and quasi-static inputs, each with its own learned parameters. For each month $t\in\{1,\ldots,12\}$, the dynamic encoder $\psi_{\mathrm{dyn}}$ maps
$X_{b,y,t}^{\mathrm{dyn}}$ to a monthly basin token $u_{b,y,t}$. A learned
month embedding $p_t\in\mathbb{R}^{d}$ is added to identify the token's
position in the annual sequence, following standard positional-encoding
practice~\citep{Vaswani2017Attention}. A Transformer with four attention heads
and GELU feed-forward blocks then models its temporal relationships across the 12 months.
The contextualized monthly tokens are aggregated to produce the annual dynamic
representation:

\begin{equation}
\begin{aligned}
u_{b,y,t}
&=
\psi_{\mathrm{dyn}}
\left(
X_{b,y,t}^{\mathrm{dyn}},
M_b
\right),
\\
\left\{
\widetilde{u}_{b,y,t}
\right\}_{t=1}^{12}
&=
\mathrm{Transformer}
\left(
\left\{
u_{b,y,t}+p_t
\right\}_{t=1}^{12}
\right),
\\
h_{b,y}^{\mathrm{dyn}}
&=
\mathrm{MLP}_{\mathrm{dyn}}
\left(
\sum_{t=1}^{12}
\beta_{b,y,t}\,
\widetilde{u}_{b,y,t}
\right).
\end{aligned}
\label{eq:monthly_branch}
\end{equation}
Here, $\widetilde{u}_{b,y,t}$ is the contextualized token for month $t$. The weight $\beta_{b,y,t}$ is produced by a learned
attention-pooling operation and normalized across the 12 months such that
$\sum_{t=1}^{12}\beta_{b,y,t}=1$. The resulting vector
$h_{b,y}^{\mathrm{dyn}}\in\mathbb{R}^{d}$ summarizes the monthly
hydroclimatic inputs for the basin-year.

The quasi-static input $X_{b,y}^{\mathrm{sta}}$ is represented by a single
multi-channel tensor for each basin and therefore does not require
temporal processing. A separate mask-aware spatial encoder $\psi_{\mathrm{sta}}$
processes its channels jointly with an MLP to produce the quasi-static
embedding:

\begin{equation}
h_{b,y}^{\mathrm{sta}}
=
\mathrm{MLP}_{\mathrm{sta}}
\left(
\psi_{\mathrm{sta}}
\left(
X_{b,y}^{\mathrm{sta}},
M_b
\right)
\right).
\label{eq:static_branch}
\end{equation}
Here, $h_{b,y}^{\mathrm{sta}}\in\mathbb{R}^{d}$ summarizes the static and
annually varying physiographic inputs for basin $b$ in year $y$.

\vspace{0.1cm}

\exphead{Fusion and Metadata Conditioning} The dynamic and quasi-static representations are fused into a single spatial
representation using learned branch-attention weights:

\begin{equation}
\begin{aligned}
\alpha_{b,y}^{\ell}
&=
\operatorname{softmax}_{\ell\in\{\mathrm{dyn},\mathrm{sta}\}}
\left(
W_{\alpha} h_{b,y}^{\ell}
\right),
\\
h_{b,y}^{\textrm{fused}}
&=
\sum_{\ell\in\{\mathrm{dyn},\mathrm{sta}\}}
\alpha_{b,y}^{\ell} h_{b,y}^{\ell}.
\end{aligned}
\label{eq:branch_attention}
\end{equation}
Here, $W_{\alpha}$ is a learned scoring layer, and the weights
$\alpha_{b,y}^{\mathrm{dyn}}$ and $\alpha_{b,y}^{\mathrm{sta}}$ are normalized
to sum to one. The resulting vector $h_{b,y}^{\textrm{fused}}\in\mathbb{R}^{d}$ combines the
spatial information learned from the static and dynamic input modalities.

Since all the basins have been mapped to the same $H\times W$ grid, their rescaled inputs do not directly convey physical basin size. Basin2Vec therefore uses an auxiliary prediction head $q_a$ to estimate the standardized log drainage area from the fused spatial representation by minimizing the following smoothed L1 loss:

\begin{equation}
\begin{aligned}
\mathcal{L}_{\mathrm{area}}
&=
\frac{1}{B}
\sum_{i=1}^{B}
\ell\left(\hat{a}_i-a_i\right),\\
\ell(e)
&=
\begin{cases}
\tfrac{1}{2}e^2, & |e|<1,\\
|e|-\tfrac{1}{2}, & |e|\geq 1.
\end{cases}
\end{aligned}
\label{eq:area_loss}
\end{equation}
Here, $B$ is the mini-batch size, $a_i$ is the standardized log drainage area of the basin associated with sample $i$, $\hat{a}_i=q_a(h^{\textrm{fused}}_i)$ is its log drainage area estimated by the prediction head $q_a$, and $e_i=\hat{a}_i-a_i$ is the prediction error. The loss is quadratic for small errors and linear for large errors, reducing sensitivity to extreme differences in basin area. Applying this objective to the fused representation $h^{\textrm{fused}}$ encourages the spatial encoder to retain information about the physical basin scale.

Finally, basin-level metadata $X_b^{\textrm{meta}}$, including drainage area, basin geometry, and effective patch resolution, are also incorporated into the embedding through gated conditioning.
An MLP initially maps the standardized metadata to the embedding dimension, while
a dimension-wise gate $g$ controls their contribution to the fused spatial
representation:

\begin{equation}
\begin{aligned}
g_{b,y}
&=
\sigma
\left(
W_g
\left[
h^{\textrm{fused}}_{b,y};
MLP_{\textrm{meta}}(X_b^{\textrm{meta}})
\right]
\right),
\\
v_{b,y}
&=
\gamma
\left(
\operatorname{LN}
\left(
h^{\textrm{fused}}_{b,y}
+
g_{b,y}\odot MLP_{\textrm{meta}}(X_b^{\textrm{meta}})
\right)
\right),
\\
z_{b,y}
&=
\operatorname{Normalize}
\left(
q_{\phi}(v_{b,y})
\right).
\end{aligned}
\label{eq:metadata_projection}
\end{equation}
Here, $[\cdot;\cdot]$ denotes concatenation, $\sigma$ is the sigmoid function,
and $g_{b,y}\in[0,1]^d$ is the metadata gate. The gated metadata are added to
$h^{\textrm{fused}}_{b,y}$, after which layer normalization and another MLP $\gamma$ is applied to produce the metadata-conditioned representation $v_{b,y}$. The projection head $q_{\phi}$
maps this representation to the embedding space, and $\ell_2$ normalization
produces the final basin-year embedding $z_{b,y}\in\mathbb{R}^{d}$.

\subsection{Relation-Aware Contrastive Objective}
\label{sec:relation_contrastive}

\begin{figure}[!htbp]
\centering
\includegraphics[
width=\linewidth,
trim={0.8cm 4cm 1cm 1.9cm},
clip
]{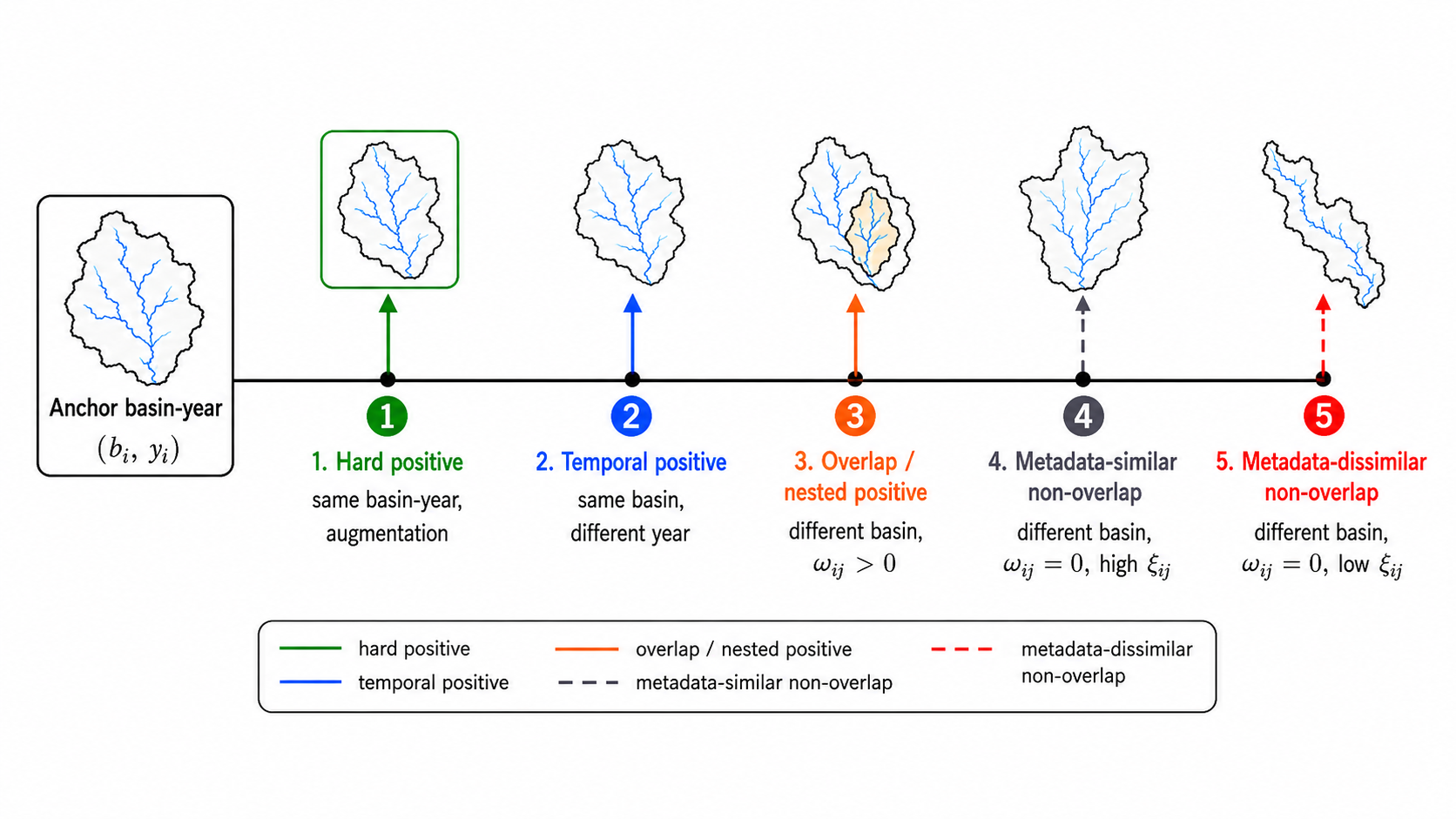}
\captionsetup{skip=2pt}
\caption{Basin relations used by the relation-aware contrastive objective.}
\label{fig:basin_relations}

\end{figure}

Standard contrastive learning treats augmented views of the same entity as
positive examples and all other entities as negative examples. This assumption is insufficient
for basin-year entities that could be related through basin
identity, temporal proximity, or spatial relationship. To overcome this limitation, Basin2Vec
considers three positive relations between basin-year entities: (1) a \textit{hard positive} between two
augmented views of the same basin-year entity, (2) a \textit{temporal positive} between two basin-year entities of the same basin, and (3) a \textit{spatial overlap positive} between entities with overlapping or nested basin boundaries. Other entity pairs without spatial
overlap remain negative examples, but their repulsive strength is adjusted using basin-metadata similarity to be described below. Figure~\ref{fig:basin_relations} summarizes
these relations.

For a mini-batch of $B$ basin-year entities, two independently augmented views
of each entity are generated and encoded into $N=2B$ normalized embeddings
$\{z_i\}_{i=1}^{N}$. Details of the augmentation procedure are provided in
Appendix~A.1. Each embedding is associated with a basin ID $b_i$, year $y_i$,
and basin metadata $X^{\textrm{meta}}_{b_i}$. The generated embeddings form an anchor set $\mathcal{A}=\{1,\ldots,N\}$ for computing contrastive loss.

For each anchor $i \in \mathcal{A}$, we define the following relations:

\begin{equation}
\begin{aligned}
\mathcal{P}_i^{\mathrm{hard}}
&=
\left\{
j :
b_i=b_j,\;
y_i=y_j,\;
i\neq j
\right\},
\\
\mathcal{P}_i^{\mathrm{temp}}
&=
\left\{
j :
b_i=b_j,\;
y_i\neq y_j
\right\},
\\
\mathcal{P}_i^{\mathrm{ov}}
&=
\left\{
j :
b_i\neq b_j,\;
\omega_{ij}>0
\right\},
\\
\mathcal{N}_i^{\mathrm{neg}}
&=
\left\{
j :
b_i\neq b_j,\;
\omega_{ij}=0
\right\}.
\end{aligned}
\label{eq:relation_sets}
\end{equation}
Here, $\mathcal{P}_i^{\mathrm{hard}}$ contains the other augmented view of the
same basin-year, $\mathcal{P}_i^{\mathrm{temp}}$ contains other years of the
same basin, $\mathcal{P}_i^{\mathrm{ov}}$ contains spatially overlapping or nested
basins, while $\mathcal{N}_i^{\mathrm{neg}}$ contains different-basin entities with no spatial overlap. We quantify the spatial overlap between embeddings $i$ and $j$ using the
Szymkiewicz--Simpson coefficient~\citep{szymkiewicz1934overlap}:
$$\omega_{ij}=|G_i\cap G_j|/\min(|G_i|,|G_j|),$$ where $G_i$ and $G_j$ denote their corresponding
basin polygons and $|\cdot|$ denotes polygon area. This coefficient reaches one when the smaller basin is fully nested within the larger, making it suitable for both overlapping and nested relations. 

Based on the relation sets defined above, our relation-aware contrastive loss for anchor $i$ is given by:

\begin{equation}
\mathcal{L}_{i}^{\mathrm{con}}
=
-
\sum_{j\neq i}
q_{ij}
\log
\frac{
d_{ij}\exp(s_{ij}/\tau)
}{
\sum_{k\neq i}
d_{ik}\exp(s_{ik}/\tau)
},
\label{eq:contrastive_loss}
\end{equation}
where $s_{ij}=z_i^\top z_j$ is the cosine similarity between two normalized
embeddings, $q_{ij}$ measures the attraction (positive) strength between embeddings, $d_{ij}$ measures the repulsive (negative) strength, while $\tau$ is a hyperparameter.

The attraction (positive) strength between embeddings $i$ and $j$ is computed as follows:

\begin{equation}
q_{ij}
=
\frac{w_{ij}^{+}}
{\sum_{k\neq i}w_{ik}^{+}},
\qquad
j\neq i.
\label{eq:positive_target_distribution}
\end{equation}
where

\begin{equation}
w_{ij}^{+}
=
\mathbf{1}_{j\in\mathcal{P}_i^{\mathrm{hard}}}
+
\lambda_t\kappa_{ij}
\mathbf{1}_{j\in\mathcal{P}_i^{\mathrm{temp}}}
+
\lambda_o\omega_{ij}\kappa_{ij}
\mathbf{1}_{j\in\mathcal{P}_i^{\mathrm{ov}}},
\label{eq:positive_weight}
\end{equation}
Here, $\lambda_t$ and $\lambda_o$ control the contributions due to temporal and spatial
overlap relations. The temporal factor
$\kappa_{ij}=\exp(-|y_i-y_j|/\tau_y)$ decreases as the
separation between years increases, with $\tau_y$ controlling the decay rate. This factor is applied to both temporal and spatial overlap relations to ensure that
basin-year entities observed far apart in time have lower attraction 
strengths. In this formulation, $q_{ij}$ is nonzero only for embedding pairs with positive relations. Since every basin-year entity has two augmented views, each anchor must have at least one hard positive.

%Pairs with $w_{ij}^{+}=0$ still contribute through the contrastive denominator. 
To avoid excessive repulsion between physiographically similar basins, Basin2Vec assigns
each pair the weight

\begin{equation}
d_{ij}
=
\begin{cases}
1, & w_{ij}^{+}>0,\\
1-\lambda_m\xi_{ij}, & w_{ij}^{+}=0,
\end{cases}
\label{eq:denominator_weight}
\end{equation}
where $\lambda_m\in[0,1)$ controls the strength of metadata-based downweighting
and $\xi_{ij}\in[0,1]$ denotes the metadata similarity between the
corresponding basins. The computation of $\xi_{ij}$ is described in
Appendix~A.1. Positive pairs retain unit weight because their attraction is already
controlled by $w_{ij}^{+}$ and $q_{ij}$. For non-positive pairs, greater metadata similarity produces a smaller $d_{ij}$ and therefore weaker repulsion. Metadata similarity is not treated as a positive relation since  physiographic similarity does not necessarily imply similar
hydrological behavior. The contrastive loss given in Equation~\eqref{eq:contrastive_loss} pulls each anchor
toward its hard, temporal, and overlap positives while reducing unnecessary repulsion from metadata-similar non-overlapping basins.

The contrastive denominator provides relative negative pressure but does not
directly bound the similarity of individual non-overlapping pairs. Basin2Vec
therefore applies the following negative-margin loss

\begin{equation}
\mathcal{L}_{\mathrm{neg}}
=
\frac{1}{|\mathcal{N}_{\mathrm{neg}}|}
\sum_{(i,j)\in\mathcal{N}_{\mathrm{neg}}}
\left[
\max(0,s_{ij}-\mu_{ij})
\right]^2,
\label{eq:negative_margin_loss}
\end{equation}
where
$\mathcal{N}_{\mathrm{neg}}
=
\{(i,j): i\in\mathcal{A},\;
j\in\mathcal{N}_i^{\mathrm{neg}}\}$
contains all non-overlapping negative pairs, and
$\mu_{ij}
=
\mu_{\min}
+
(\mu_{\max}-\mu_{\min})\xi_{ij}$
is the metadata-dependent margin. Metadata-similar negative pairs receive a
larger margin and may remain closer in the embedding space, whereas
metadata-dissimilar pairs receive a stricter margin.

Altogether, the overall pre-training objective is

\begin{equation}
\mathcal{L}
=
\frac{1}{N}
\sum_{i\in\mathcal{A}}
\mathcal{L}_{i}^{\mathrm{con}}
+
\lambda_n\mathcal{L}_{\mathrm{neg}}
+
\lambda_a\mathcal{L}_{\mathrm{area}},
\label{eq:total_loss}
\end{equation}
where the coefficients
$\lambda_n$ and $\lambda_a$ control the contributions of the negative-margin
and area-preservation terms, respectively. Note that
$\mathcal{L}_{i}^{\mathrm{con}}$ organizes the embedding space using known
basin relations, $\mathcal{L}_{\mathrm{neg}}$ limits excessive similarity
between non-overlapping basins, while $\mathcal{L}_{\mathrm{area}}$ encourages
the encoder to retain information about physical basin scale.

\subsection{Training and Embedding Generation}
\label{sec:training_embedding_generation}

A basin-aware sampler selects the basins for each mini-batch and draws $K_{\mathrm{yr}}$ years per
basin, to ensure the same number of temporal positives for each basin within a mini-batch. After training, the unaugmented input for each original basin-year entity is encoded to obtain its final embedding $z_{b,y}$.

Finally, to obtain a fixed descriptor for each basin, we take the average annual embeddings for each basin and $\ell_2$-normalize the result:
\begin{equation}
\bar{z}_b
=
\operatorname{Normalize}
\left(
\frac{1}{|\mathcal{Y}_b|}
\sum_{y\in\mathcal{Y}_b} z_{b,y}
\right),
\label{eq:basin_embedding}
\end{equation}
where $\mathcal{Y}_b$ denotes the years available for basin $b$. This facilitates a fair downstream comparison since some EO baselines have different temporal coverages, i.e., do not provide embeddings for a common set of years
as Basin2Vec. 

\section{Experimental Setup}
\label{sec:experiments}

We evaluate Basin2Vec through three research questions (RQs) covering
intrinsic embedding structure, temporal generalization, and zero-shot
spatial transfer.

\subsection{RQ1: Intrinsic Embedding Structure}
\label{sec:rq1}

RQ1 tests whether the learned representations (i) remain diverse rather than
collapsing into a low-dimensional subspace and (ii) preserve metadata-defined
scale and shape relationships. For each method, we average the available embeddings per basin, apply $\ell_2$ normalization,
and evaluate the same 9,067 basins.

\paragraph{Representation diversity.}
We measure representation diversity using entropy-based effective rank~\citep{Roy2007EffectiveRank}. Given the eigenvalues
$\{\lambda_i\}_{i=1}^{d}$ of the covariance matrix of the embeddings, let
$p_i=\lambda_i/\sum_{j=1}^{d}\lambda_j$. The effective rank of the learned embeddings is

\begin{equation}
r_{\mathrm{eff}}
=
\exp\left(-\sum_{i=1}^{d}p_i\log p_i\right).
\label{eq:effective_rank}
\end{equation}
Since the embedding dimensions may differ for different methods, we report the normalized effective rank $\widetilde{r}_{\mathrm{eff}}
=r_{\mathrm{eff}}/\min(d,N-1)$, where $d$ and $N$ denote the embedding
dimensions and number of basins, respectively. Higher values indicate that the embeddings are more diverse as their variance spans a larger fraction of the available dimensions. 

\paragraph{Metadata consistency.}
The metadata incorporated into Basin2Vec include descriptors of basin scale and
geometry (Appendix~A.2), which influence basin hydrologic behavior. These
descriptors are integrated through metadata fusion, while pairwise metadata
similarity is used in the relation-aware objective. We therefore test whether metadata-similar basins are closer in the learned
embedding space than metadata-dissimilar basins.

We use standardized scale-and-shape metadata (Appendix A.2), with no
geographic-coordinate inputs, and exclude overlapping and nested pairs.
\textit{Metadata-similar pairs} are defined using each basin's nearest neighbors in
Euclidean metadata space. \textit{Metadata-dissimilar pairs} are selected as the most
distant pairs within a randomly sampled pool of non-overlapping candidates.
The same pair sets are evaluated for every method. We compare the
$\ell_2$-distance distributions of normalized basin embeddings and summarize
their separation as
$\Delta_{\mathrm{meta}}=\mu_{\mathrm{dissimilar}}-
\mu_{\mathrm{similar}}$, where the two terms denote the respective mean
pairwise distances. A larger positive $\Delta_{\mathrm{meta}}$ indicates
clearer preservation of metadata-defined basin relationships.

\subsection*{RQ2--RQ3: Downstream Streamflow Generalization}
\label{sec:downstream_task}

\exphead{Task description}
RQ2 and RQ3 evaluate basin representations as static conditioning vectors for
daily streamflow prediction. For basin $b$ and target day $t$, the downstream
model predicts streamflow $Q_{b,t}$ from an input sequence of meteorological forcings $\mathbf{x}^{\mathrm{met}}_{b,t}$ and a fixed basin
descriptor $c_b$:

\begin{equation}
\widehat{Q}_{b,t}
=
F_{\eta}
\left(
[\mathbf{x}^{\mathrm{met}}_{b,t-L+1},\ldots,
\mathbf{x}^{\mathrm{met}}_{b,t}],c_b
\right).
\label{eq:downstream_prediction}
\end{equation}
where $L\in\{120,365\}$ is the forcing-window length. %We consider same-day (lead-zero) prediction with $L\in\{120,365\}$.
Meteorological forcings are available through day $t$, but no observed
streamflow is provided as input.

\exphead{Data description}
Experiments use CAMELS-US basins with Daymet meteorological forcings and USGS
daily streamflow~\citep{Newman2015CAMELS,Addor2017CAMELS}. The dynamic inputs
are precipitation, shortwave radiation, maximum and minimum temperature, and
vapor pressure. Discharge is converted to basin-mean runoff (mm/day); the
27 CAMELS attributes include drainage area (\texttt{area\_gages2}).
The main benchmark compares basin descriptors obtained from
Basin2Vec, AlphaEarth, TESSERA, SatCLIP, and CAMELS attributes.

Because the learned representations differ in temporal coverage and spatial
sampling, we construct one fixed descriptor per basin using all samples
available from each source. Basin2Vec uses the $\ell_2$-normalized,
year-averaged representation $\bar{z}_b$ from
Eq.~\eqref{eq:basin_embedding}. For the EO baselines, embeddings are
averaged over the available years and sampled locations within each basin and
then $\ell_2$-normalized. This avoids selecting an arbitrary source-dependent
year and provides a time-invariant descriptor comparable to the CAMELS
attributes. All basin descriptors remain frozen during downstream training.

\exphead{Evaluation protocol}
For RQ2, \emph{temporal generalization} is evaluated on the same 671 basins
across disjoint periods. Data from 1990--2009 are used for model development,
including early stopping, while 2010--2014 is held out for evaluation
~\citep{Gauch2021CAMELS,Fang2022DataSynergy}.

For RQ3, \emph{zero-shot spatial transfer} is evaluated using 6-fold
ungauged-basin cross-validation. In each fold, held-out basins are excluded
from supervised training and early stopping. The remaining basins use
1990--2004 for training and 2005--2014 for validation, while the held-out
basins are evaluated over 1990--2014
~\citep{Sivapalan2003PUB,Bloschl2013PUB,Kratzert2019PUB}.

The downstream predictor $F_{\eta}$ is instantiated as
LSTM~\citep{Hochreiter1997LSTM}, TFT~\citep{Lim2021TFT},
TimeXer~\citep{Wang2024TimeXer}, Timer-XL~\citep{Liu2025TimerXL}, or
iTransformer~\citep{Liu2024iTransformer}. Within each architecture, the
dynamic inputs, conditioning mechanism, and training protocol are held fixed
across representation sources; only the descriptor input width changes to
match the source dimension. All inputs, targets, and conditioning vectors are standardized using training data only.

We additionally compare Basin2Vec with KG-SSL~\citep{ghosh2022robust}, a
hydrology-specific self-supervised method, using its unsupervised
reconstruction-and-contrastive setting without static-attribute supervision.
KG-SSL is trained and its descriptors are constructed from 1990--2009
meteorology and observed streamflow, then frozen for 2010--2014 evaluation.
The comparison uses the matched downstream LSTM with $L=365$ and lead-zero
prediction. KG-SSL is excluded from RQ3 because its descriptor construction
requires streamflow from the held-out basin.

Models are optimized using MSE and AdamW with validation-based early stopping
and gradient clipping. Each configuration is evaluated using three seeds
($42$, $123$, and $456$). For each seed, NSE~\citep{Nash1970NSE} and
KGE~\citep{Gupta2009KGE} are computed per basin and summarized by their median
across basins. We report the mean and standard deviation of these medians over
the three seeds; higher values indicate better performance, with $1$
representing perfect agreement.

\textit{Complete hyperparameter and implementation details are provided in the
appendix.}

\begin{table}[!htbp]
\centering
\caption{ Temporal generalization results. Values are median basin-level NSE and KGE. \textbf{Bold} and \underline{underlining} indicate the best and second-best conditioning sources, respectively. }
\label{tab:temporal_h1_five_sources}

\begingroup
\fontsize{7.4pt}{8.0pt}\selectfont
\setlength{\tabcolsep}{1.8pt}
\renewcommand{\arraystretch}{1.08}

\begin{adjustbox}{max width=\textwidth}
\begin{adjustbox}{max width=\linewidth}
\begin{tabular}{@{}l|ccccc|ccccc@{}}
\toprule

\multicolumn{11}{c}{$L=365$} \\
\midrule

Model
& \multicolumn{5}{c|}{Median NSE $\uparrow$}
& \multicolumn{5}{c}{Median KGE $\uparrow$} \\
\cmidrule(lr){2-6}\cmidrule(lr){7-11}

& Basin2Vec & AlphaEarth & CAMELS Attr. & TESSERA & SatCLIP
& Basin2Vec & AlphaEarth & CAMELS Attr. & TESSERA & SatCLIP \\
\midrule

LSTM
& $\mathbf{0.782\!\pm\!0.009}$
& $0.641\!\pm\!0.017$
& $\underline{0.747\!\pm\!0.010}$
& $0.662\!\pm\!0.025$
& $0.719\!\pm\!0.018$
& $\mathbf{0.786\!\pm\!0.011}$
& $0.572\!\pm\!0.013$
& $\underline{0.756\!\pm\!0.013}$
& $0.626\!\pm\!0.022$
& $0.688\!\pm\!0.031$ \\

TFT
& $\mathbf{0.753\!\pm\!0.011}$
& $\underline{0.741\!\pm\!0.026}$
& $0.717\!\pm\!0.022$
& $0.570\!\pm\!0.025$
& $0.638\!\pm\!0.026$
& $\mathbf{0.768\!\pm\!0.019}$
& $\underline{0.722\!\pm\!0.029}$
& $0.714\!\pm\!0.030$
& $0.514\!\pm\!0.033$
& $0.566\!\pm\!0.026$ \\

TimeXer
& $\mathbf{0.651\!\pm\!0.018}$
& $\underline{0.630\!\pm\!0.012}$
& $0.502\!\pm\!0.019$
& $0.573\!\pm\!0.033$
& $0.603\!\pm\!0.015$
& $\mathbf{0.691\!\pm\!0.009}$
& $\underline{0.579\!\pm\!0.023}$
& $0.446\!\pm\!0.030$
& $0.517\!\pm\!0.023$
& $0.571\!\pm\!0.033$ \\

Timer-XL
& $\mathbf{0.730\!\pm\!0.019}$
& $\underline{0.674\!\pm\!0.028}$
& $0.591\!\pm\!0.026$
& $0.627\!\pm\!0.019$
& $0.402\!\pm\!0.021$
& $\mathbf{0.708\!\pm\!0.022}$
& $0.585\!\pm\!0.010$
& $0.536\!\pm\!0.029$
& $\underline{0.635\!\pm\!0.021}$
& $0.346\!\pm\!0.016$ \\

iTrans.
& $\mathbf{0.627\!\pm\!0.022}$
& $0.559\!\pm\!0.020$
& $0.555\!\pm\!0.022$
& $\underline{0.576\!\pm\!0.015}$
& $0.464\!\pm\!0.017$
& $\mathbf{0.582\!\pm\!0.026}$
& $0.566\!\pm\!0.024$
& $0.533\!\pm\!0.020$
& $\underline{0.579\!\pm\!0.028}$
& $0.450\!\pm\!0.022$ \\

\midrule
\multicolumn{11}{c}{$L=120$} \\
\midrule

LSTM
& $\mathbf{0.754\!\pm\!0.014}$
& $0.619\!\pm\!0.015$
& $\underline{0.737\!\pm\!0.013}$
& $0.638\!\pm\!0.016$
& $0.704\!\pm\!0.030$
& $\mathbf{0.775\!\pm\!0.009}$
& $0.563\!\pm\!0.023$
& $\underline{0.741\!\pm\!0.031}$
& $0.592\!\pm\!0.031$
& $0.705\!\pm\!0.025$ \\

TFT
& $\mathbf{0.737\!\pm\!0.019}$
& $\underline{0.719\!\pm\!0.020}$
& $0.712\!\pm\!0.012$
& $0.551\!\pm\!0.029$
& $0.668\!\pm\!0.026$
& $\mathbf{0.744\!\pm\!0.020}$
& $\underline{0.719\!\pm\!0.032}$
& $0.668\!\pm\!0.030$
& $0.479\!\pm\!0.014$
& $0.600\!\pm\!0.020$ \\

TimeXer
& $\mathbf{0.677\!\pm\!0.011}$
& $\underline{0.675\!\pm\!0.010}$
& $0.645\!\pm\!0.026$
& $0.592\!\pm\!0.014$
& $0.530\!\pm\!0.028$
& $\mathbf{0.679\!\pm\!0.025}$
& $0.625\!\pm\!0.016$
& $\underline{0.675\!\pm\!0.011}$
& $0.563\!\pm\!0.028$
& $0.492\!\pm\!0.010$ \\

Timer-XL
& $\mathbf{0.705\!\pm\!0.018}$
& $\underline{0.650\!\pm\!0.026}$
& $0.529\!\pm\!0.021$
& $0.602\!\pm\!0.019$
& $0.566\!\pm\!0.016$
& $\mathbf{0.693\!\pm\!0.014}$
& $\underline{0.580\!\pm\!0.016}$
& $0.440\!\pm\!0.029$
& $0.540\!\pm\!0.027$
& $0.533\!\pm\!0.025$ \\

iTrans.
& $\mathbf{0.594\!\pm\!0.019}$
& $0.551\!\pm\!0.015$
& $0.581\!\pm\!0.028$
& $\underline{0.584\!\pm\!0.017}$
& $0.398\!\pm\!0.025$
& $\mathbf{0.580\!\pm\!0.009}$
& $0.545\!\pm\!0.010$
& $\underline{0.554\!\pm\!0.012}$
& $0.535\!\pm\!0.031$
& $0.349\!\pm\!0.024$ \\

\bottomrule
\end{tabular}
\end{adjustbox}
\end{adjustbox}

\endgroup

\end{table}

\begin{figure}[!htbp]
    \centering
    \includegraphics[width=\columnwidth]
    {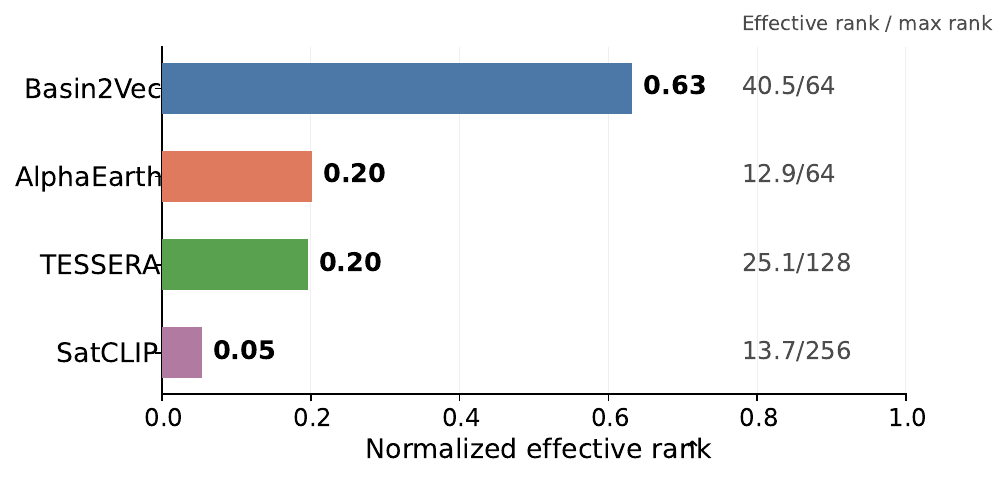}
    \captionsetup{skip=2pt}
    \caption{\textbf{Representation diversity.}
    Normalized effective rank across 9,067 basins. Higher values indicate greater representation diversity.}
    \label{fig:effective_rank}
    
\end{figure}

\begin{figure}[!htbp]
    \centering
    \includegraphics[width=\linewidth, trim={0.1cm 0.30cm 0.1cm 0.25cm}, clip]
    {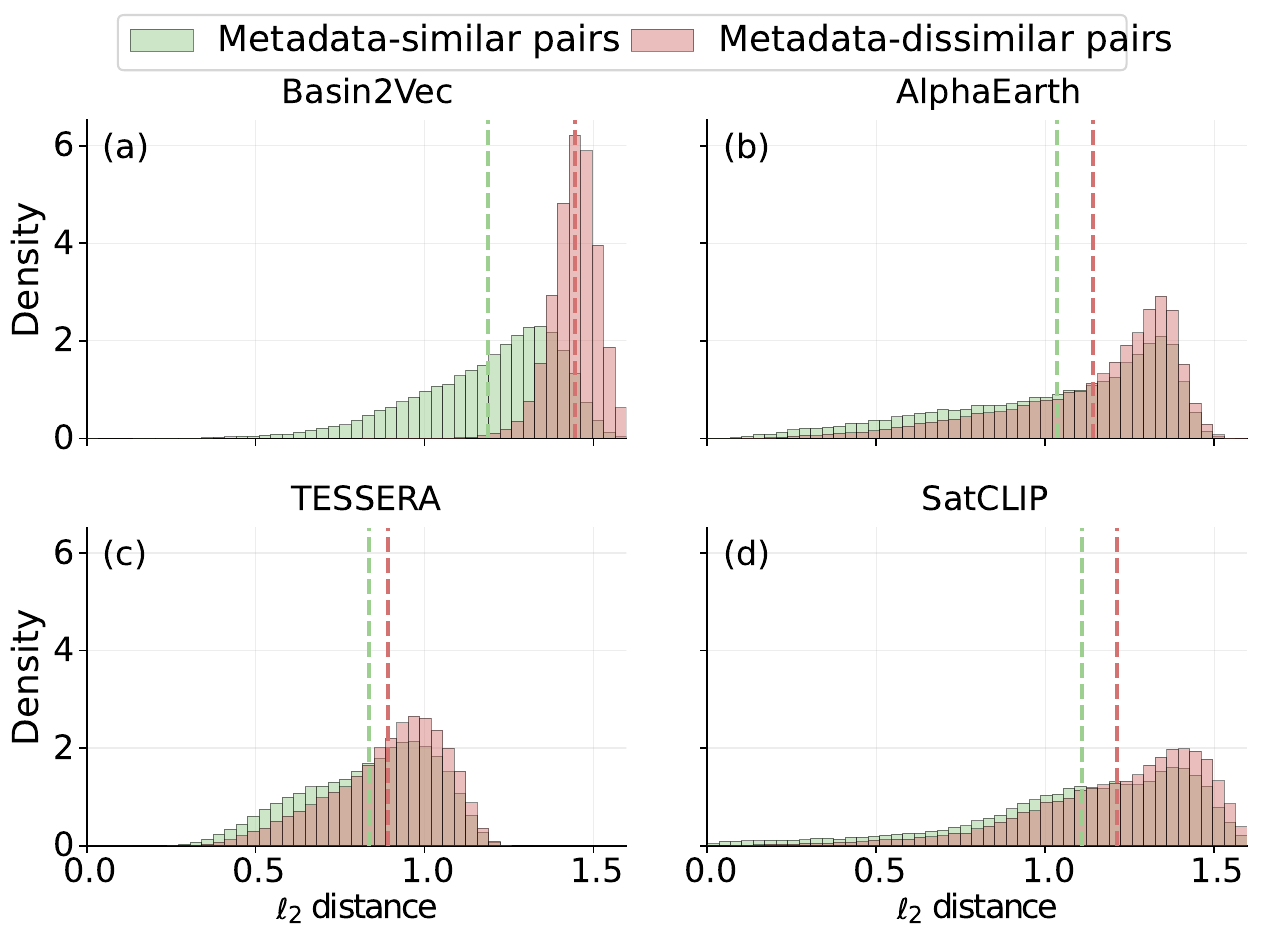}
    \captionsetup{skip=2pt}
    \caption{\textbf{Metadata consistency.}
    $\ell_2$-distance distributions for metadata-similar and metadata-dissimilar non-overlapping pairs. Dashed lines denote means.}
    \label{fig:metadata_consistency}
    
\end{figure}

\begin{table}[!htbp]
\centering
\caption{ Zero-shot spatial-transfer results for $L=365$. Each value is the median per-basin NSE or KGE on held-out basins. \textbf{Bold} and \underline{underlining} indicate the best and second-best conditioning sources, respectively. }
\label{tab:spatial_h1_five_sources}

\begingroup
\fontsize{7.4pt}{8.0pt}\selectfont
\setlength{\tabcolsep}{1.8pt}
\renewcommand{\arraystretch}{1.08}

\begin{adjustbox}{max width=\textwidth}
\begin{adjustbox}{max width=\linewidth}
\begin{tabular}{@{}l|ccccc|ccccc@{}}
\toprule
Model
& \multicolumn{5}{c|}{Median NSE $\uparrow$}
& \multicolumn{5}{c}{Median KGE $\uparrow$} \\
\cmidrule(lr){2-6}\cmidrule(lr){7-11}
& Basin2Vec & AlphaEarth & CAMELS Attr. & TESSERA & SatCLIP
& Basin2Vec & AlphaEarth & CAMELS Attr. & TESSERA & SatCLIP \\
\midrule

LSTM
& $\mathbf{0.460\!\pm\!0.050}$
& $0.079\!\pm\!0.021$
& $0.071\!\pm\!0.018$
& $0.092\!\pm\!0.017$
& $\underline{0.132\!\pm\!0.022}$
& $\mathbf{0.408\!\pm\!0.001}$
& $-0.084\!\pm\!0.025$
& $-0.098\!\pm\!0.027$
& $\underline{-0.049\!\pm\!0.020}$
& $-0.072\!\pm\!0.024$ \\

TFT
& $\mathbf{0.524\!\pm\!0.019}$
& $0.132\!\pm\!0.017$
& $\underline{0.147\!\pm\!0.015}$
& $0.127\!\pm\!0.018$
& $0.119\!\pm\!0.021$
& $\mathbf{0.427\!\pm\!0.022}$
& $-0.077\!\pm\!0.028$
& $-0.071\!\pm\!0.025$
& $\underline{-0.055\!\pm\!0.019}$
& $-0.100\!\pm\!0.023$ \\

TimeXer
& $\mathbf{0.491\!\pm\!0.016}$
& $\underline{0.126\!\pm\!0.020}$
& $0.065\!\pm\!0.018$
& $0.065\!\pm\!0.021$
& $0.106\!\pm\!0.017$
& $\mathbf{0.431\!\pm\!0.018}$
& $\underline{-0.050\!\pm\!0.022}$
& $-0.080\!\pm\!0.019$
& $-0.055\!\pm\!0.020$
& $-0.066\!\pm\!0.024$ \\

Timer-XL
& $\mathbf{0.498\!\pm\!0.017}$
& $0.136\!\pm\!0.024$
& $0.092\!\pm\!0.019$
& $0.022\!\pm\!0.016$
& $\underline{0.140\!\pm\!0.021}$
& $\mathbf{0.436\!\pm\!0.020}$
& $-0.038\!\pm\!0.021$
& $-0.060\!\pm\!0.023$
& $-0.114\!\pm\!0.029$
& $\underline{-0.023\!\pm\!0.018}$ \\

iTrans.
& $\mathbf{0.465\!\pm\!0.021}$
& $\underline{0.151\!\pm\!0.018}$
& $0.007\!\pm\!0.014$
& $0.085\!\pm\!0.017$
& $0.058\!\pm\!0.020$
& $\mathbf{0.384\!\pm\!0.024}$
& $\underline{-0.039\!\pm\!0.020}$
& $-0.093\!\pm\!0.026$
& $-0.098\!\pm\!0.024$
& $-0.114\!\pm\!0.028$ \\

\bottomrule
\end{tabular}
\end{adjustbox}
\end{adjustbox}

\endgroup

\end{table}

\begin{table}[!htbp]
\centering
\caption{Component ablations of Basin2Vec using LSTM with $L=365$. \textbf{Bold} and \underline{underlining} denote the best and second-best
results.}
\label{tab:expanded_ablation_lstm_l365}

\begingroup
\fontsize{7.4pt}{8.2pt}\selectfont
\setlength{\tabcolsep}{3pt}
\renewcommand{\arraystretch}{1.28}

\begin{adjustbox}{max width=\linewidth}
\begin{tabular}{lll|cc|cc}
\hline
Category
& Variant
& Ablation setting
& \multicolumn{2}{c|}{Temporal}
& \multicolumn{2}{c}{Spatial} \\
\cline{4-7}
&
&
& NSE $\uparrow$
& KGE $\uparrow$
& NSE $\uparrow$
& KGE $\uparrow$ \\
\hline

Full model
& \textbf{Basin2Vec}
& All components
& $\mathbf{0.782 \pm 0.009}$
& $\mathbf{0.786 \pm 0.011}$
& $\mathbf{0.460 \pm 0.050}$
& $\mathbf{0.408 \pm 0.001}$ \\

\hline

\multirow{5}{*}{\shortstack[l]{Scale-aware\\encoder}}
& w/o monthly branch
& $z^{\mathrm{dyn}}_{b,y}$ removed
& $0.649 \pm 0.016$
& $0.621 \pm 0.031$
& $0.423 \pm 0.021$
& $0.388 \pm 0.032$ \\

& w/o quasi-static branch
& $z^{\mathrm{sta}}_{b,y}$ removed
& $0.635 \pm 0.015$
& $0.623 \pm 0.014$
& $0.404 \pm 0.013$
& $0.397 \pm 0.007$ \\

& w/o metadata conditioning
& $\mathrm{MLP}_{\mathrm{meta}}()$ and $g_{b,y}$ removed
& $0.634 \pm 0.022$
& $0.615 \pm 0.017$
& $0.064 \pm 0.044$
& $-0.059 \pm 0.044$ \\

& w/o gated metadata fusion
& gated fusion $\rightarrow$ concat.\,+\,MLP
& $0.651 \pm 0.032$
& $0.620 \pm 0.059$
& $\underline{0.424 \pm 0.029}$
& $\underline{0.399 \pm 0.042}$ \\

& w/o area regularization
& $\lambda_a=0$
& $0.601 \pm 0.007$
& $0.571 \pm 0.001$
& $0.412 \pm 0.013$
& $0.389 \pm 0.014$ \\

\hline

\multirow{6}{*}{\shortstack[l]{Relation-aware\\objective}}
& hard positive only
& $w^+_{ij}=\mathbf{1}_{j\in\mathcal{P}^{\mathrm{hard}}_i},
\;d_{ij}=1$
& $\underline{0.670 \pm 0.009}$
& $\underline{0.647 \pm 0.021}$
& $0.372 \pm 0.010$
& $0.332 \pm 0.021$ \\

& w/o temporal positives
& $\lambda_t=0$
& $0.621 \pm 0.024$
& $0.628 \pm 0.056$
& $0.418 \pm 0.030$
& $0.398 \pm 0.011$ \\

& w/o overlap/nesting positives
& $\lambda_o=0$
& $0.633 \pm 0.012$
& $0.589 \pm 0.001$
& $0.408 \pm 0.017$
& $0.389 \pm 0.017$ \\

& w/o metadata-weighted denominator
& $d_{ij}=1$
& $0.638 \pm 0.028$
& $0.608 \pm 0.035$
& $0.384 \pm 0.017$
& $0.330 \pm 0.008$ \\

& metadata as positive
& $w^+_{ij}\mathrel{+}=\lambda_p\xi_{ij}$
& $0.600 \pm 0.039$
& $0.580 \pm 0.040$
& $0.403 \pm 0.002$
& $0.390 \pm 0.004$ \\

& w/o negative-margin loss
& $\lambda_n=0$
& $0.643 \pm 0.032$
& $0.625 \pm 0.048$
& $0.409 \pm 0.024$
& $0.383 \pm 0.029$ \\

\hline
\end{tabular}
\end{adjustbox}
\endgroup

\vspace{0.8em}

\end{table}

\begin{table}[!htbp]
\centering
\caption{Comparison with KG-SSL under the matched LSTM setting with $L=365$. }
\label{tab:kgssl_temporal}
\begin{adjustbox}{max width=\linewidth}
\begin{tabular}{lcc}
\toprule
Method & Median NSE $\uparrow$ & Median KGE $\uparrow$ \\
\midrule
KG-SSL
& $0.620\!\pm\!0.018$
& $0.593\!\pm\!0.015$ \\
Basin2Vec
& $\mathbf{0.782\!\pm\!0.009}$
& $\mathbf{0.786\!\pm\!0.011}$ \\
\bottomrule
\end{tabular}
\end{adjustbox}
\end{table}

\section{Experimental Results}
\label{sec:results}

\subsection{Intrinsic Embedding Structure\texorpdfstring{\hfill\textit{(RQ1)}}{ (RQ1)}}
\label{sec:results_intrinsic}

Basin2Vec achieves the highest normalized effective rank
(Fig.~\ref{fig:effective_rank}), indicating that its variance is distributed
across a larger fraction of the available dimensions rather than concentrated
in a small subspace. All methods place metadata-dissimilar pairs farther apart
on average than metadata-similar pairs, but Basin2Vec exhibits the largest
separation $\Delta_{\mathrm{meta}}$, while the baseline distributions overlap
more substantially (Fig.~\ref{fig:metadata_consistency}). These results show
that Basin2Vec produces a diverse, non-collapsed representation that more
clearly preserves basin scale and geometry relationships.

\subsection{Temporal Generalization\texorpdfstring{\hfill\textit{(RQ2)}}{ (RQ2)}}
\label{sec:results_temporal}

\begin{figure}[!htbp]
\centering
\includegraphics[width=\linewidth, trim={0.1cm 0.3cm 0.1cm 0.2cm}, clip]{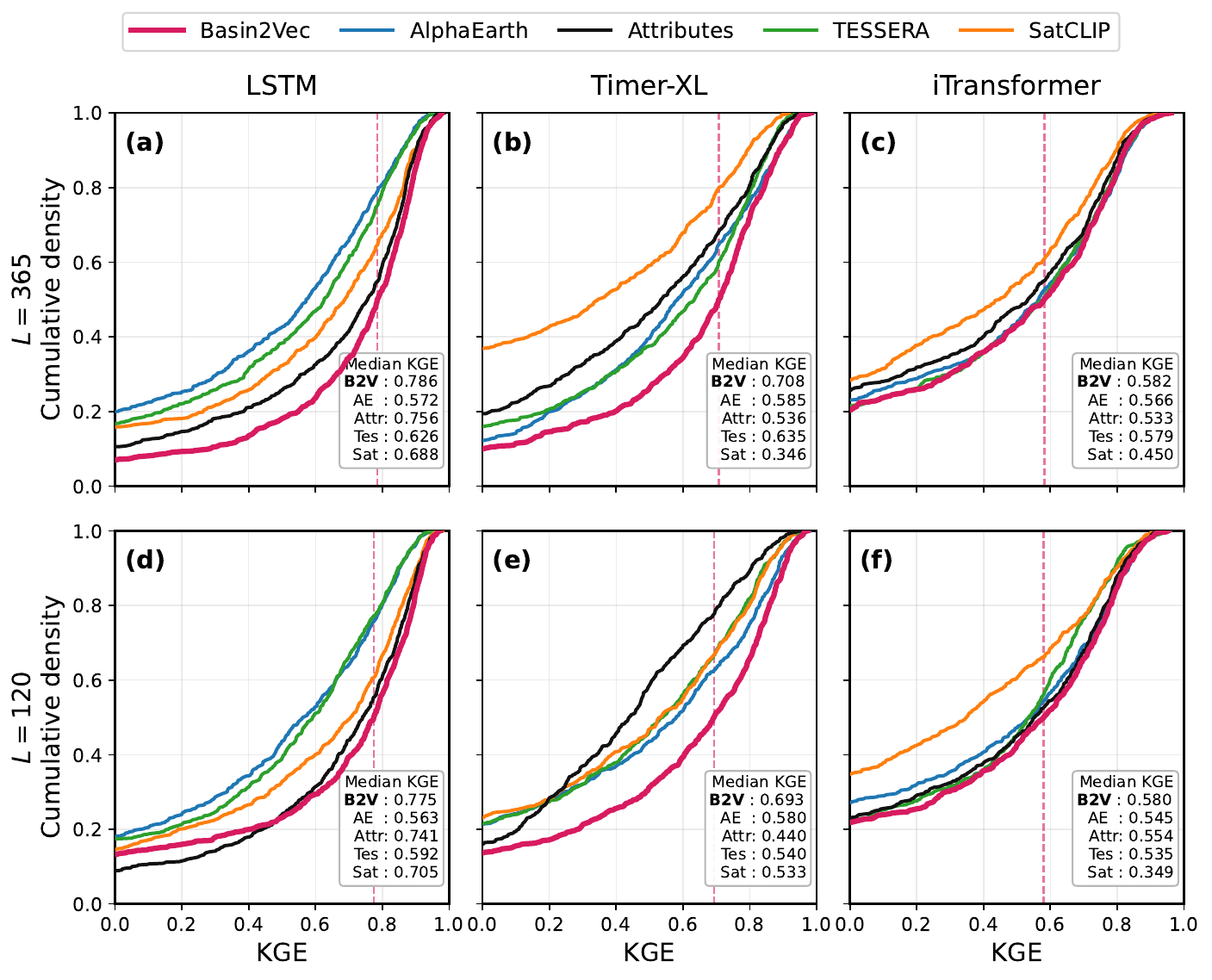}
\captionsetup{skip=2pt}
\caption{ Per-basin KGE CDFs for temporal generalization using selected downstream architectures.}
\label{fig:temporal_cdf}

\end{figure}

Table~\ref{tab:temporal_h1_five_sources} summarizes the temporal generalization performance of the embedding methods across five downstream architectures. Basin2Vec ranks first in both NSE and KGE for every architecture at forcing-window length $L\in\{120,365\}$. Its consistent gains across recurrent, attention-based, and transformer-style predictors indicate that the benefit is not specific to one architecture or forcing-window length. The right-shifted KGE distributions in Figure~\ref{fig:temporal_cdf} further show that the gains extend across basins.

\paragraph{Comparison with hydrology-specific SSL.}
With the same LSTM and $L=365$, Basin2Vec increases median NSE from $0.620$ to
$0.782$ and median KGE from $0.593$ to $0.786$ over
KG-SSL~\citep{ghosh2022robust} (Table~\ref{tab:kgssl_temporal}). Although KG-SSL
uses streamflow, it learns from lumped one-year sequences over 671 basins,
making its descriptors more dependent on the observed flow regime. Basin2Vec
instead learns from distributed inputs across 9,067 basins and enforces
consistency across years, producing more stable descriptors for temporal
transfer.

% {\color{red} The reported results are based on LSTM architecture with L=365. KG-SSL uses streamflow to train their embeddings, yet their performance is worse. Why?}

\subsection{Zero-Shot Spatial Transfer\texorpdfstring{\hfill\textit{(RQ3)}}{ (RQ3)}}
\label{sec:results_spatial}

\begin{figure}[!htbp]
\centering
\includegraphics[width=\linewidth, trim={0.25cm 0.33cm 0.1cm 0.1cm}, clip]{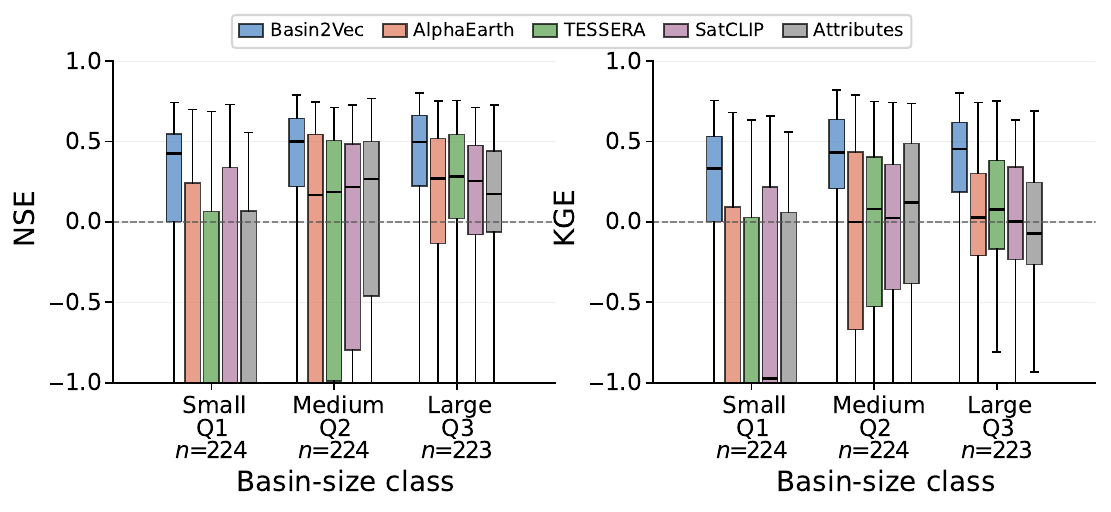}
\captionsetup{skip=2pt}
\caption{
Spatial-split performance across basin-size classes.}
\Description{
Boxplots comparing per-basin NSE and KGE across small, medium, and large basin-size classes for Basin2Vec, AlphaEarth, TESSERA, SatCLIP, and CAMELS attributes.
}
\label{fig:area_class_spatial}

\end{figure}

Table~\ref{tab:spatial_h1_five_sources} shows that Basin2Vec ranks first in NSE
and KGE for every downstream architecture. Its median NSE exceeds the strongest alternative for each architecture by
$0.314$--$0.377$, and it is the only method with
consistently positive median KGE, demonstrating substantially stronger transfer
to unseen basins. We also examine the relative performance of the methods across different basin sizes. Figure~\ref{fig:area_class_spatial} further shows that Basin2Vec maintains the strongest NSE and KGE distributions for small, medium, and large basins. Thus, the gains are not confined to a particular basin-size range.

% 
% \subsection{Ablation Study}
% \label{sec:ablation}

% Table~\ref{tab:expanded_ablation_lstm_l365} isolates the contributions of the
% principal Basin2Vec design choices under temporal and spatial generalization.
% 
% \paragraph{Encoder and Data Fusion.}
% The monthly, quasi-static, and metadata pathways all contribute to
% generalization, with metadata being particularly important for unseen basins.
% Gated fusion also outperforms direct concatenation, supporting adaptive
% integration of raster and basin-level information.
% 
% \paragraph{Basin-aware relations.}
% Temporal and overlap/nesting positives each provide useful supervision beyond
% augmented views of the same basin-year, confirming the value of incorporating
% hydrologically meaningful relations during pretraining.
% 
% \paragraph{Metadata-aware contrastive design.}
% Metadata similarity is more effective for weighting non-positive pairs than for
% defining additional positives. This design modulates negative pressure without
% forcing distinct but physiographically similar basins together.
% 
% \paragraph{Auxiliary objectives.}
% The negative-margin loss complements the contrastive denominator by limiting
% excessive similarity among non-overlapping basins, while area regularization
% provides a scale-aware constraint. Removing either objective degrades
% generalization.

% Hyperparameter sensitivity is examined in Appendix~A.5.

% 

\subsection{Ablation Study}
\label{sec:ablation}

We remove or replace individual components of the scale-aware encoder and
relation-aware objective and evaluate the resulting variants under temporal and
spatial generalization. Table~\ref{tab:expanded_ablation_lstm_l365} shows that
the complete Basin2Vec model achieves the best result for every metric. \textit{Scale-aware encoder.} Removing either input branch degrades
generalization, while removing metadata conditioning causes the largest spatial
decline and negative median KGE. Gated metadata fusion outperforms
concatenation, and area regularization provides particularly strong temporal
gains. \textit{Relation-aware objective.} The full objective outperforms every relation
ablation. Hard positives alone rank second temporally but transfer poorly to
unseen basins. Removing temporal or overlap/nesting positives,
metadata-weighted negatives, or the negative-margin loss reduces performance.
Treating metadata as a positive relation also performs worse, supporting its
use to moderate negative repulsion rather than enforce attraction.

\section{Conclusion}
\label{sec:conclusion}

We presented Basin2Vec, a self-supervised framework for learning basin-centric
representations from gridded hydroclimatic data without observed streamflow.
Basin2Vec combines a scale-aware encoder for spatiotemporal inputs and metadata
with a relation-aware contrastive objective that models hydrologic relationships
across space and time. The resulting embeddings are diverse and hydrologically
structured. Across architectures and forcing windows, Basin2Vec outperforms
predefined attributes and general-purpose EO embeddings in temporal
generalization and zero-shot transfer to unseen basins. Ablations validate its
main design choices. These results show that explicitly modeling basin structure
and relationships yields reusable representations for hydrologic prediction.

\section*{Acknowledgments}
Pang-Ning Tan acknowledges partial support from the U.S. National Science Foundation under grant IIS-2453100. A. Pouyan Nejadhashemi acknowledges support from the USDA National Institute of Food and Agriculture (NIFA) through Hatch project 1019654 and from the Institute of Water Research at Michigan State University.

\clearpage
\bibliographystyle{colm2025_conference}
\bibliography{main}
\clearpage
\noindent
The appendix is organized as follows:
Sec.~A.1 provides additional Basin2Vec implementation details;
Sec.~A.2 describes the pretraining and baseline datasets;
Sec.~A.3 provides downstream training and evaluation details;
Sec.~A.4 reports extended results for RQ1--RQ3; and
Sec.~A.5 reports grouped metadata and objective ablations from Table~4.

% -------------------- Appendix numbering --------------------
\appendix

\renewcommand{\thefigure}{A.\arabic{figure}}
\renewcommand{\thetable}{A.\arabic{table}}
\renewcommand{\theequation}{A.\arabic{equation}}
\renewcommand{\thealgorithm}{A.\arabic{algorithm}}

\setcounter{figure}{0}
\setcounter{table}{0}
\setcounter{equation}{0}
\setcounter{algorithm}{0}

\section{Appendix}
\label{app:appendix}

\subsection{Additional Basin2Vec Implementation Details}
\label{app:basin2vec_details}

This section complements the Basin2Vec formulation in Sec.~3 of the main
paper with the implementation details needed to reproduce pretraining and
embedding extraction. We specify the metadata-similarity kernel, the exact
parameterization of metadata-aware denominator weights and margins,
mask-preserving augmentations, basin-aware distributed batching, and the
embedding-extraction procedure.

\paragraph{Metadata similarity.}
Let $\widetilde{X}_{b_i}^{\textrm{meta}}$ and
$\widetilde{X}_{b_j}^{\textrm{meta}}$ denote the standardized metadata vectors
for the basins associated with samples $i$ and $j$, respectively. The metadata
fields are listed in Table~\ref{tab:basin2vec_metadata}. For metadata dimension
$M_{\mathrm{meta}}$, we compute the normalized Euclidean distance
\begin{equation}
\delta^{m}_{ij}
=
\frac{
\left\|
\widetilde{X}_{b_i}^{\textrm{meta}}
-
\widetilde{X}_{b_j}^{\textrm{meta}}
\right\|_2
}{\sqrt{M_{\mathrm{meta}}}}.
\label{eq:app_meta_dist}
\end{equation}
The metadata similarity $\xi_{ij}$ used by the denominator weight in
Eq.~(11) of the main paper and the negative-margin loss in Eq.~(12) is then
computed with a Gaussian kernel:
\begin{equation}
\xi_{ij}
=
\exp\left(
-\frac{(\delta^{m}_{ij})^2}{2\tau_m^2}
\right),
\qquad
\xi_{ij}\in[0,1],
\label{eq:app_meta_sim}
\end{equation}
where $\tau_m$ is the metadata-similarity bandwidth. This normalization keeps
the distance scale from increasing mechanically with the number of metadata
fields.

\paragraph{Denominator-weight parameterization.}
For a non-positive pair, Eq.~(11) of the main paper defines
$d_{ij}=1-\lambda_m\xi_{ij}$. In the implementation, the same weight is
parameterized using the minimum negative weight
$d_{\min}=1-\lambda_m$:
\begin{equation}
d_{ij}
=
\begin{cases}
1, & w^+_{ij}>0,\\
d_{\min}+(1-d_{\min})(1-\xi_{ij}), & w^+_{ij}=0.
\end{cases}
\label{eq:app_denom}
\end{equation}
The two expressions are algebraically equivalent. With
$\lambda_m=0.80$, the implementation uses $d_{\min}=0.20$, so
metadata-similar non-positive pairs exert weaker repulsion but are never
removed from the contrastive denominator. Positive pairs retain unit
denominator weight because their attraction is already determined by
$w^+_{ij}$ and $q_{ij}$.

\paragraph{Metadata-dependent negative margin.}
For each non-overlapping negative pair, the margin in Eq.~(12) of the main
paper is
\begin{equation}
\mu_{ij}
=
\mu_{\min}
+
(\mu_{\max}-\mu_{\min})\xi_{ij}.
\label{eq:app_margin}
\end{equation}
We use $\mu_{\min}=0.20$ and $\mu_{\max}=0.60$. Thus,
metadata-similar negative pairs may retain a higher cosine similarity, whereas
metadata-dissimilar negative pairs are subject to a stricter margin.

\paragraph{Mask-preserving augmentations.}
For every basin-year entity, two views are independently generated from
$X_{b,y}^{\mathrm{dyn}}$ and $X_{b,y}^{\mathrm{sta}}$ using weak
perturbations that preserve the basin boundary. Continuous non-temperature
variables are multiplicatively scaled by $1+\epsilon$, where
$\epsilon\sim\mathcal{N}(0,0.04^2)$, while standardized temperature channels
receive an additive shift with
$\epsilon\sim\mathcal{N}(0,0.03^2)$. Gaussian noise is applied with standard
deviation $0.03$. Spatial dropout uses one spatial mask shared across all
12 months so that it does not introduce artificial month-specific gaps.
For the dynamic input, individual months are dropped with probability $0.08$,
and seasonal-block dropout removes a contiguous block of one to three months
with probability $0.08$. Whole-variable dropout is applied with probability
$0.03$. Categorical channels are not perturbed by noise, scaling, or spatial
dropout and may undergo only whole-variable dropout. After augmentation,
continuous values are clipped to $[-10,10]$, and the basin mask $M_b$ is
reapplied. We do not use cropping, rotation, or month permutation because they
would alter basin geometry or seasonal ordering.

\begin{table}[!htbp]
\centering
\scriptsize
\setlength{\tabcolsep}{4pt}
\renewcommand{\arraystretch}{1.04}
\caption{Default Basin2Vec pretraining and augmentation settings.}
\label{tab:augmentation_settings}
\begin{adjustbox}{max width=\linewidth}
\begin{tabular}{lc}
\toprule
Setting & Value \\
\midrule
Patch size & $64\times64$ \\
Embedding dimension $d$ & $64$ \\
Monthly Transformer heads & $4$ \\
Contrastive temperature $\tau$ & $0.15$ \\
Temporal-decay scale $\tau_y$ & $8.0$ \\
Metadata bandwidth $\tau_m$ & $1.5$ \\
Temporal-positive weight $\lambda_t$ & $0.35$ \\
Overlap-positive weight $\lambda_o$ & $0.12$ \\
Metadata-denominator weight $\lambda_m$ & $0.80$ \\
Minimum negative weight $d_{\min}$ & $0.20$ \\
Negative-margin weight $\lambda_n$ & $0.05$ \\
Area-loss weight $\lambda_a$ & $0.05$ \\
Margin bounds $(\mu_{\min},\mu_{\max})$ & $(0.20,0.60)$ \\
\midrule
Gaussian noise & $\sigma=0.03$ \\
Spatial dropout & $p=0.03$ \\
Month dropout & $p=0.08$ \\
Seasonal-block dropout & $p=0.08$, length $1$--$3$ \\
Whole-variable dropout & $p=0.03$ \\
Multiplicative scaling & $\sigma=0.04$ \\
Temperature shift & $\sigma=0.03$ \\
\bottomrule
\end{tabular}
\end{adjustbox}
\end{table}

\paragraph{Objective coefficients and ablation scope.}
Table~\ref{tab:augmentation_settings} specifies the reference objective coefficients, including $\lambda_t=0.35$, $\lambda_o=0.12$, $\lambda_m=0.80$, and $\lambda_n=\lambda_a=0.05$. The component-removal comparisons in Table~4 change the stated objective term while retaining the reference settings for other terms. Such comparisons assess component contributions; they do not constitute a coefficient search or establish an optimal setting. A validation-based search protocol, candidate-selection rule, and coefficient-selection metric are not documented here, so no claim of validation-optimal coefficients is made.

\paragraph{Basin-aware distributed batching.}
Each local minibatch is formed by sampling $B_b$ distinct basins and
$K_{\mathrm{yr}}$ years per selected basin, giving
$B_{\mathrm{loc}}=B_bK_{\mathrm{yr}}$ basin-year entities. This construction
ensures that each selected basin contributes temporal positives. Under
distributed training with $R$ workers, embeddings, basin IDs, years, metadata,
overlap information, and area targets are gathered across workers before the
loss is evaluated. The effective global batch therefore contains
$B=RB_{\mathrm{loc}}$ basin-year entities and $N=2B$ augmented embeddings.
All hard, temporal, overlap/nesting, and non-overlapping relations are formed
over this global batch rather than separately on each worker.

\begin{algorithm}[t]
\caption{Basin2Vec pretraining and basin-level embedding extraction}
\label{alg:basin2vec}
\footnotesize
\begin{algorithmic}[1]

\Require Basin-year data
$\mathcal{D}=\{(X^{\mathrm{dyn}}_{b,y},
X^{\mathrm{sta}}_{b,y},M_b,X_b^{\textrm{meta}},a_b)\}$,
where $a_b$ is standardized log drainage area
\Require Overlap strengths $\omega_{ij}$; encoder $f_\theta$;
projection head $q_\phi$; area head $q_a$; augmentations
$\mathcal{T}_{\mathrm{hydro}}$
\Require Sampling parameters $B_b,K_{\mathrm{yr}}$ and objective parameters
$\tau,\tau_y,\tau_m,\lambda_t,\lambda_o,\lambda_m,
\lambda_n,\lambda_a,\mu_{\min},\mu_{\max}$
\Ensure Basin-level descriptors $\{\bar{z}_b\}$

\For{each pretraining epoch}
    \For{each distributed minibatch}
        \State Sample $B_b$ basins and $K_{\mathrm{yr}}$ years per basin
        to form $\mathcal{B}_{\mathrm{loc}}$.

        \For{each $(b,y)\in\mathcal{B}_{\mathrm{loc}}$}
            \State Generate two independent mask-preserving views with
            $\mathcal{T}_{\mathrm{hydro}}$.
            \For{$r\in\{1,2\}$}
                \State Encode the view to obtain
                $h_{b,y}^{\textrm{fused},(r)}$ and
                $v_{b,y}^{(r)}$.
                \State
                $z_{b,y}^{(r)}\gets
                \operatorname{Normalize}(q_\phi(v_{b,y}^{(r)}))$.
                \State
                $\widehat{a}_{b,y}^{(r)}\gets
                q_a(h_{b,y}^{\textrm{fused},(r)})$.
            \EndFor
        \EndFor

        \State Gather embeddings, identifiers, metadata, overlap information,
        and area predictions and targets across all workers.
        \State Construct $\mathcal{P}_i^{\mathrm{hard}}$,
        $\mathcal{P}_i^{\mathrm{temp}}$, $\mathcal{P}_i^{\mathrm{ov}}$, and
        $\mathcal{N}_i^{\mathrm{neg}}$ using Eq.~(7) of the main paper.
        \State Compute $\kappa_{ij}$ and $w^+_{ij}$ using Eq.~(10), and
        normalize $w^+_{ij}$ to obtain $q_{ij}$ using Eq.~(9).
        \State Compute $\xi_{ij}$ using Eq.~\eqref{eq:app_meta_sim},
        $d_{ij}$ using Eq.~\eqref{eq:app_denom}, and
        $\mu_{ij}$ using Eq.~\eqref{eq:app_margin}.
        \State Compute $\mathcal{L}_{\mathrm{con}}$ using Eq.~(8),
        $\mathcal{L}_{\mathrm{neg}}$ using Eq.~(12), and
        $\mathcal{L}_{\mathrm{area}}$ using Eq.~(5) of the main paper.
        \State
        $\mathcal{L}\gets
        \mathcal{L}_{\mathrm{con}}
        +\lambda_n\mathcal{L}_{\mathrm{neg}}
        +\lambda_a\mathcal{L}_{\mathrm{area}}$.
        \State Update the trainable parameters by backpropagating
        $\mathcal{L}$.
    \EndFor
\EndFor

\State Set the trained network to evaluation mode.
\For{each basin $b$}
    \State Encode every unaugmented available basin-year to obtain
    $\{z_{b,y}:y\in\mathcal{Y}_b\}$.
    \State
    $\displaystyle
    \bar{z}_b\gets
    \operatorname{Normalize}\!\left(
    \frac{1}{|\mathcal{Y}_b|}
    \sum_{y\in\mathcal{Y}_b}z_{b,y}
    \right)$.
\EndFor
\State \Return $\{\bar{z}_b\}$

\end{algorithmic}
\end{algorithm}

\paragraph{Optimization and checkpointing.}
Basin2Vec is trained with distributed data parallelism using pre-stacked,
grouped Zarr inputs. Optimization uses AdamW, mixed-precision training, and
gradient clipping. Both views share the same encoder and projection head, and
the area loss is averaged across the two views. Checkpoints store the encoder,
projection head, and area head together with the input-channel configuration,
metadata columns, normalization statistics, objective configuration, and
sampling configuration required for embedding extraction.

\paragraph{Embedding extraction.}
After pretraining, the network is placed in evaluation mode and embeddings are
computed from unaugmented inputs under no-gradient inference. During
distributed extraction, dataset indices are sharded without padding so that
each basin-year entity is evaluated exactly once. The annual embeddings are
then averaged across all available years for each basin and $\ell_2$-normalized
to obtain $\bar{z}_b$, as defined in Eq.~(14) of the main paper. Downstream
experiments use only this final basin-level descriptor.

\subsection{Dataset Details}
\label{app:data}

\subsubsection{Pretraining Data}
\label{app:pretraining_data}

\exphead{BASINGRID}
Basin2Vec is pretrained on BASINGRID, which provides basin-aligned gridded
data for 9,067 GAGES-II basins across the contiguous United States from
1985--2024~\citep{talha2026basingrid,Falcone2017GAGES}. BASINGRID combines
daily meteorological forcings, annually varying variables, and distributed
static attributes. Each variable is stored in Zarr format with dimensions
organized by basin, year, time, channel, height, and width, together with a
binary basin mask and basin-window metadata.

All source rasters are first harmonized to the common Daymet reference grid
(approximately 1~km). Continuous variables are aligned using bilinear
resampling, whereas categorical variables use nearest-neighbor resampling.
For each basin, a padded extraction window is then resampled to a
$64\times64$ patch and masked to exclude cells outside the basin boundary.
The fixed array size ensures consistent model inputs and cross-variable
alignment, but not a common physical resolution: the ground area represented
by one patch pixel varies with basin extent. Basin-window dimensions and
effective patch-pixel size are therefore retained in the metadata.

For Basin2Vec pretraining, the six daily meteorological variables are
aggregated to monthly resolution, producing the 12-month dynamic input
$X_{b,y}^{\mathrm{dyn}}$. Annual PET and land cover are matched to year $y$,
while the remaining static variables are reused across years; together they
form the quasi-static input $X_{b,y}^{\mathrm{sta}}$. Table~\ref{tab:basin2vec_data_sources}
summarizes the BASINGRID variables and their use in Basin2Vec.

\begin{table}[!htbp]
\centering
\scriptsize
\setlength{\tabcolsep}{3.2pt}
\renewcommand{\arraystretch}{1.05}
\caption{BASINGRID variables used for Basin2Vec pretraining. Native daily
variables are aggregated to monthly resolution before pretraining. Source
definitions follow BASINGRID~\citep{talha2026basingrid}.}
\label{tab:basin2vec_data_sources}
\begin{tabularx}{\textwidth}{@{}lllp{0.19\textwidth}lX@{}}
\toprule
\textbf{Group} & \textbf{Variable} & \textbf{BASINGRID type}
& \textbf{Basin2Vec input} & \textbf{Unit} & \textbf{Source} \\
\midrule
Hydroclimate
& Precipitation & Daily & Monthly dynamic & mm/day & Daymet \\
& Minimum temperature & Daily & Monthly dynamic & $^\circ$C & Daymet \\
& Maximum temperature & Daily & Monthly dynamic & $^\circ$C & Daymet \\
& Downward shortwave radiation & Daily & Monthly dynamic & W/m$^2$ & Daymet \\
& Snow water equivalent & Daily & Monthly dynamic & kg/m$^2$ & Daymet \\
& Vapor pressure & Daily & Monthly dynamic & Pa & Daymet \\
& Potential evapotranspiration & Annual & Annual quasi-static
& mm/year & Hargreaves method using CW3E temperatures \\
\midrule
Topography
& Elevation & Static & Quasi-static & m & USGS 3DEP \\
\midrule
Soil
& Clay fraction & Static, multi-depth & Quasi-static & g/kg & SoilGrids \\
& Silt fraction & Static, multi-depth & Quasi-static & g/kg & SoilGrids \\
& Sand fraction & Static, multi-depth & Quasi-static & g/kg & SoilGrids \\
& Bulk density & Static, multi-depth & Quasi-static & cg/cm$^3$ & SoilGrids \\
\midrule
Hydrogeology
& Near-surface permeability & Static & Quasi-static & m$^2$ & GLHYMPS V2.0 \\
& Hydraulic conductivity & Static & Quasi-static & m/s & GLHYMPS V2.0 \\
\midrule
Land cover
& Land-cover classification & Annual & Annual quasi-static & -- & NLCD \\
\midrule
Agriculture
& Tile drainage & Static & Quasi-static & -- & AgTile-US \\
\bottomrule
\end{tabularx}
\end{table}

\begin{figure}[!htbp]
\centering
\includegraphics[width=\textwidth]
{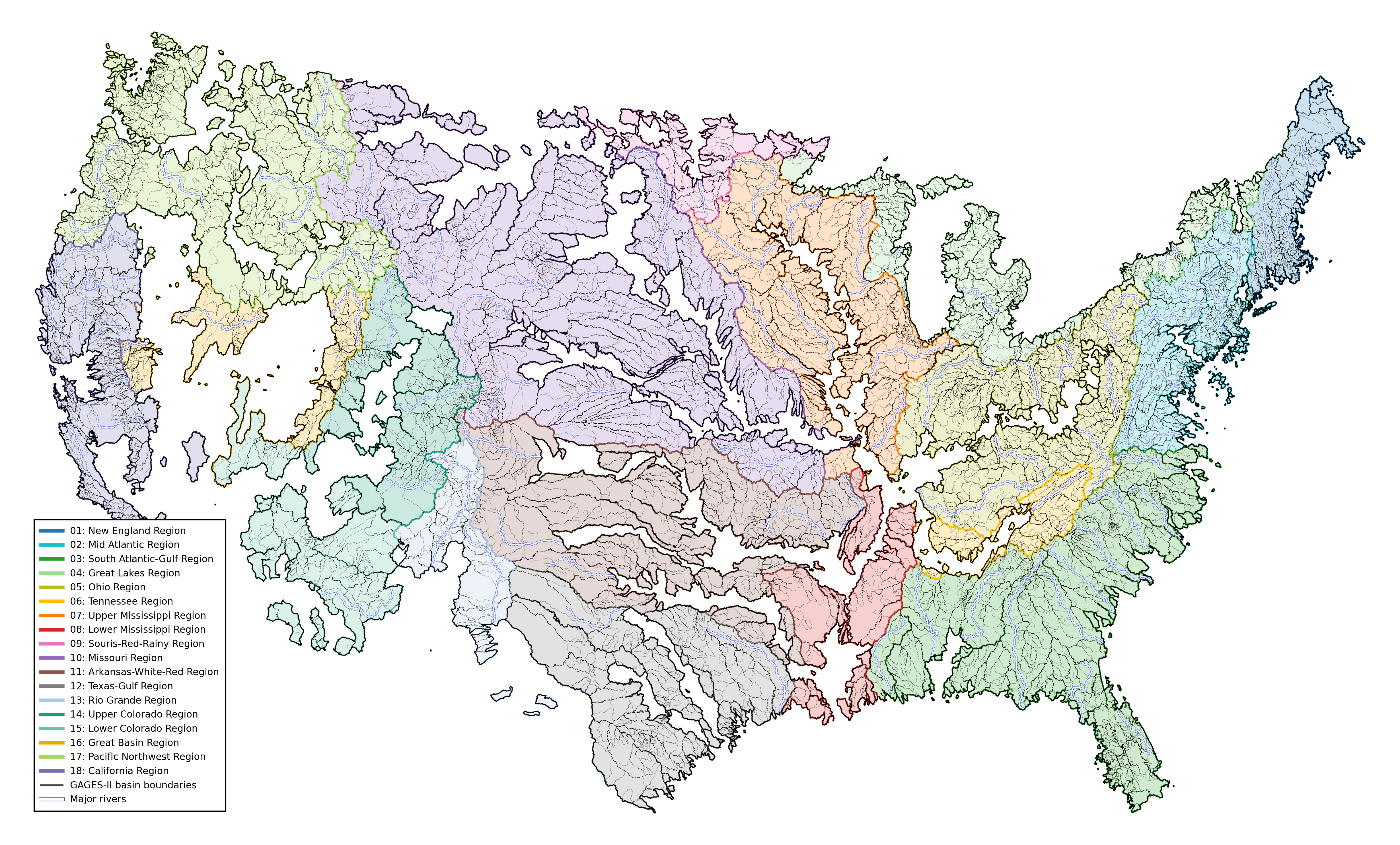}
\caption{Spatial coverage of the 9,067 BASINGRID basins used for Basin2Vec
pretraining. Colors denote HUC-2 hydrologic regions, and major rivers are
shown in light blue.}
\Description{Map of the contiguous United States showing 9,067 GAGES-II basin
boundaries colored by HUC-2 hydrologic region, with major rivers overlaid.}
\label{fig:conus_huc2_basin_coverage}
\end{figure}
\exphead{Basin metadata}
The BASINGRID metadata file contains raw basin geometry, patch-scale
information, and derived standardized variables for all 9,067 basins.
Basin2Vec constructs the metadata vector
$X_b^{\textrm{meta}}\in\mathbb{R}^{18}$ from the standardized fields listed in
Table~\ref{tab:basin2vec_metadata}. Right-skewed scale, geometry, boundary, and
resolution quantities are log-transformed during preprocessing, after which
each feature is standardized across the pretraining basins. Basin identifiers,
centroid coordinates, and topology counts are not included. The input schema
has 18 fields, of which 15 vary across basins; the final three fields are
constant zero. Thus, the metadata provides 15 nonconstant input features,
not 18 independently varying quantities.

The same metadata vector is incorporated through gated conditioning in the
scale-aware encoder and is used to compute metadata similarity for
non-overlapping pairs in the relation-aware objective. The pairwise overlap
coefficient $\omega_{ij}$ used to construct overlap and nesting positives is
computed separately from basin polygons. The three overlap-summary fields are
retained in the metadata schema but are constant zero in the supplied training
file and therefore provide no basin-dependent information for conditioning
or metadata similarity. They are distinct from the pair-specific polygon
overlap coefficients: zero overlap-summary inputs do not imply zero overlap
relations or remove overlap/nesting positives from the objective.

\begin{table}[!htbp]
\centering
\scriptsize
\setlength{\tabcolsep}{4pt}
\renewcommand{\arraystretch}{1.03}
\caption{Standardized basin metadata fields supplied to Basin2Vec. The final
three overlap-summary fields are constant zero in the training metadata file;
the remaining 15 fields are nonconstant.}
\label{tab:basin2vec_metadata}

\begin{tabularx}{\columnwidth}{@{}p{0.57\columnwidth}X@{}}
\toprule
\textbf{Variable} & \textbf{Quantity} \\
\midrule
\varname{log_area_km2_z}
& Drainage area \\

\varname{log_perimeter_km_z}
& Basin perimeter \\

\varname{log_equivalent_diameter_km_z}
& Equivalent diameter \\

\varname{log_bbox_width_km_z}
& Bounding-box width \\

\varname{log_bbox_height_km_z}
& Bounding-box height \\

\varname{log_bbox_diagonal_km_z}
& Bounding-box diagonal \\

\varname{log_bbox_aspect_ratio_z}
& Bounding-box aspect ratio \\

\varname{bbox_fill_fraction_z}
& Basin-to-bounding-box area ratio \\

\varname{compactness_z}
& Basin compactness \\

\varname{solidity_z}
& Basin solidity \\

\varname{log_shape_index_z}
& Basin shape index \\

\varname{log_vertex_density_per_km_z}
& Boundary vertex density \\

\varname{log_patch_pixel_width_km_z}
& Effective patch-pixel width \\

\varname{log_patch_pixel_height_km_z}
& Effective patch-pixel height \\

\varname{log_patch_pixel_area_km2_z}
& Effective patch-pixel area \\

\varname{log_overlap_degree_z}
& Number of overlapping basins \\

\varname{max_overlap_strength_z}
& Maximum overlap strength \\

\varname{mean_overlap_strength_z}
& Mean overlap strength \\
\bottomrule
\end{tabularx}
\end{table}

\paragraph{Metadata used for RQ1.}
The metadata-consistency analysis in RQ1 uses a restricted scale-and-shape
subset comprising drainage area, perimeter, equivalent diameter, bounding-box
width, height, area, diagonal, aspect ratio, and fill fraction, together with
compactness, shape index, and solidity. These 12 quantities are standardized
before pair construction. Centroid coordinates are not part of the encoder metadata and are not added
for RQ1; geographic distance is therefore not an explicit feature in this
evaluation. Effective
patch-resolution, boundary-complexity, and overlap-summary variables are also
excluded. Spatially overlapping and nested basin pairs are removed before
selecting metadata-similar and metadata-dissimilar pairs.

\subsubsection{Baseline Data}
\label{app:baseline_data}

\exphead{CAMELS-US}
The downstream experiments use the 671 CAMELS-US basins with Daymet
meteorological forcings and USGS daily streamflow
observations~\citep{Newman2015CAMELS,Addor2017CAMELS}. The meteorological
inputs are precipitation, shortwave radiation, maximum temperature, minimum
temperature, and vapor pressure. Observed discharge is converted to basin-mean
runoff in mm/day using the corresponding CAMELS drainage area.

The hand-crafted descriptor baseline uses 27 CAMELS attributes describing
climate, topography, vegetation, soil, and geology. Table~\ref{tab:camels_attributes}
lists the exact attributes. Categorical identifiers and attributes with missing
values are excluded. The resulting vectors are standardized using statistics
computed from the supervised training basins only.

\begin{table}[!htbp]
\centering
\scriptsize
\setlength{\tabcolsep}{3.5pt}
\renewcommand{\arraystretch}{1.05}
\caption{CAMELS-US attributes used as the hand-crafted basin descriptor
baseline. The descriptor contains 27 variables.}
\label{tab:camels_attributes}

\begin{tabularx}{\columnwidth}{@{}p{0.20\columnwidth}X@{}}
\toprule
\textbf{Group} & \textbf{Attributes} \\
\midrule

Climate
& \texttt{p\_mean}, \texttt{pet\_mean}, \texttt{p\_seasonality},
\texttt{frac\_snow}, \texttt{aridity}, \texttt{high\_prec\_freq},
\texttt{high\_prec\_dur}, \texttt{low\_prec\_freq}, and
\texttt{low\_prec\_dur} \\

Topography
& \texttt{elev\_mean}, \texttt{slope\_mean}, and
\texttt{area\_gages2} \\

Vegetation
& \texttt{frac\_forest}, \texttt{lai\_max}, \texttt{lai\_diff},
\texttt{gvf\_max}, and \texttt{gvf\_diff} \\

Soil
& \texttt{soil\_depth\_pelletier}, \texttt{soil\_depth\_statsgo},
\texttt{soil\_porosity}, \texttt{soil\_conductivity},
\texttt{max\_water\_content}, \texttt{sand\_frac},
\texttt{silt\_frac}, and \texttt{clay\_frac} \\

Geology
& \texttt{carbonate\_rocks\_frac} and
\texttt{geol\_permeability} \\

\bottomrule
\end{tabularx}
\end{table}

\exphead{Earth-observation embeddings}
We additionally compare Basin2Vec with three general-purpose geospatial
representations: AlphaEarth, TESSERA, and
SatCLIP~\citep{Brown2025AlphaEarth,Feng2025TESSERA,Klemmer2025SatCLIP}.
Because these representations are defined over pixels or geographic
coordinates, they are spatially aggregated over each CAMELS basin to construct
a fixed basin descriptor.

For AlphaEarth, the 64-dimensional annual embedding images are averaged over
each basin polygon using Earth Engine region reduction. The resulting annual
descriptors from 2017--2024 are then averaged to obtain one vector per basin.
For TESSERA, 128-dimensional embeddings from the 2024 product are sampled at
basin-interior locations and averaged across those locations. For SatCLIP,
the time-invariant 256-dimensional location embeddings are evaluated at the
same basin-interior locations and spatially averaged. Each aggregated EO
descriptor is $\ell_2$-normalized and kept fixed during downstream training.

Thus, every representation source provides one time-invariant conditioning
vector per basin: a year-averaged Basin2Vec embedding, a CAMELS attribute
vector, or a spatially aggregated EO embedding. None of these descriptors is
updated by the downstream forecasting models.

\subsection{Downstream Training and Evaluation Details}
\label{app:downstream_training}

\exphead{Task and preprocessing}
Each basin representation is used as a fixed conditioning vector and is not
fine-tuned during downstream training. The predictor receives Daymet
precipitation, shortwave radiation, maximum temperature, minimum temperature,
and vapor pressure over a forcing window of
$L\in\{120,365\}$ days and predicts streamflow for the final day of the window.
Meteorological observations through the target day are available, corresponding
to lead-zero prediction with horizon $H=1$. Past streamflow is never provided
as an input.

Dynamic forcings and streamflow targets are standardized using statistics
computed from the corresponding supervised training split. Predictions are
transformed back to the original runoff scale before evaluation. Static
conditioning vectors are also standardized using the appropriate training
scope: all 671 basins for temporal generalization, because the basin set is
shared across periods, and only the non-test basins within each spatial fold.

\exphead{Evaluation splits}
Table~\ref{tab:app_downstream_protocols} summarizes the temporal and spatial
evaluation protocols. In RQ2, the same basins appear during model development
and evaluation, but the time periods are disjoint. Data from 1990--2009 are
used for training and early stopping, while 2010--2014 is held out for final
evaluation~\citep{Gauch2021CAMELS,Fang2022DataSynergy}.

In RQ3, the 671 basins are partitioned into six folds without HUC-2
stratification. For each fold, the held-out basins are excluded from supervised
training, validation, early stopping, and feature standardization. The
remaining basins use 1990--2004 for training and 2005--2014 for validation,
while the held-out basins are evaluated over 1990--2014. Thus, every basin is
evaluated exactly once as an unseen basin
~\citep{Sivapalan2003PUB,Bloschl2013PUB,Kratzert2019PUB}.

\begin{table}[!htbp]
\centering
\scriptsize
\setlength{\tabcolsep}{2.5pt}
\renewcommand{\arraystretch}{1.05}
\caption{Downstream streamflow-prediction protocols.}
\label{tab:app_downstream_protocols}
\begin{tabularx}{\columnwidth}{@{}p{0.27\columnwidth}X@{}}
\toprule
\textbf{Setting} & \textbf{Protocol} \\
\midrule
Task
& Daily streamflow prediction, lead $0$, horizon $H=1$ \\
Forcing windows
& $L\in\{120,365\}$ days \\
Dynamic inputs
& Five Daymet meteorological forcings through target day $t$ \\
Static input
& One fixed basin descriptor \\
Discharge input
& None \\
Temporal development
& Same 671 basins, 1990--2009 \\
Temporal evaluation
& Same 671 basins, 2010--2014 \\
Spatial evaluation
& Six-fold ungauged-basin cross-validation \\
Spatial training
& Non-test basins, 1990--2004 \\
Spatial validation
& Non-test basins, 2005--2014 \\
Spatial testing
& Held-out basins, 1990--2014 \\
Random seeds
& $42$, $123$, and $456$ \\
Metrics
& Basin-level NSE and KGE \\
\bottomrule
\end{tabularx}
\end{table}

\exphead{Descriptor conditioning}
Descriptor construction is detailed in Appendix~A.2. Basin2Vec uses the
normalized year-averaged descriptor $\bar{z}_b$, whereas the EO baselines use
spatially aggregated and normalized embeddings and CAMELS uses its predefined
attribute vector. All descriptors remain fixed during downstream training.

Each architecture uses the same conditioning mechanism for every descriptor
source. For the LSTM, the descriptor is first projected to a common hidden
dimension and concatenated with the meteorological input at every time step.
For the attention-based predictors, it is projected to the model dimension and
provided as static or exogenous context according to the corresponding
architecture. Only the input width of the descriptor projection changes across
representation sources.

\exphead{Optimization and repeatability}
All downstream models minimize mean squared error using AdamW, gradient
clipping, mixed-precision training, and validation-based early stopping.
Reduce-on-plateau scheduling decreases the learning rate by a factor of
$0.5$ after five epochs without validation improvement. The optimization
settings are given in Table~\ref{tab:app_downstream_optimization}.

Each model--descriptor--window configuration is evaluated using seeds
$42$, $123$, and $456$. For each seed, evaluation metrics are first computed
independently for every basin and summarized by their median across basins.
The reported result is the mean and standard deviation of these three
seed-specific medians.

\begin{table}[!htbp]
\centering
\scriptsize
\setlength{\tabcolsep}{2.5pt}
\renewcommand{\arraystretch}{1.05}
\caption{Downstream optimization settings.}
\label{tab:app_downstream_optimization}
\begin{adjustbox}{max width=\linewidth}
\begin{tabular}{@{}p{0.48\columnwidth}
>{\centering\arraybackslash}p{0.22\columnwidth}
>{\centering\arraybackslash}p{0.22\columnwidth}@{}}
\toprule
\textbf{Setting} & \textbf{Temporal} & \textbf{Spatial} \\
\midrule
Maximum epochs & 40 & 30 \\
Early-stopping patience & 5 & 8 \\
Batch size per GPU & 128 & 128 \\
Optimizer & AdamW & AdamW \\
Initial learning rate & $10^{-3}$ & $10^{-4}$ \\
Weight decay & $10^{-4}$ & $10^{-4}$ \\
Training loss & MSE & MSE \\
Mixed precision & Yes & Yes \\
Gradient-norm clipping & $1.0$ & $1.0$ \\
Scheduler factor & $0.5$ & $0.5$ \\
Scheduler patience & 5 & 5 \\
Minimum learning rate & $10^{-5}$ & $10^{-5}$ \\
\bottomrule
\end{tabular}
\end{adjustbox}
\end{table}

\exphead{Model configurations}
We evaluate LSTM, TFT, TimeXer, Timer-XL, and
iTransformer~\citep{Hochreiter1997LSTM,Lim2021TFT,Wang2024TimeXer,
Liu2025TimerXL,Liu2024iTransformer}. Table~\ref{tab:app_downstream_hyperparams}
reports the shared configuration used for each architecture group. Settings
that are not applicable to a particular architecture are omitted by that
implementation.

\begin{table}[!htbp]
\centering
\scriptsize
\setlength{\tabcolsep}{2.2pt}
\renewcommand{\arraystretch}{1.05}
\caption{Model hyperparameters used in downstream experiments.}
\label{tab:app_downstream_hyperparams}
\begin{tabularx}{\columnwidth}
{@{}p{0.32\columnwidth}X
>{\centering\arraybackslash}p{0.16\columnwidth}
>{\centering\arraybackslash}p{0.16\columnwidth}@{}}
\toprule
\textbf{Architecture} & \textbf{Hyperparameter}
& \textbf{Temporal} & \textbf{Spatial} \\
\midrule
LSTM
& Hidden dimension & 128 & 256 \\
& Recurrent layers & 2 & 2 \\
\midrule
Attention-based
& Model dimension & 128 & 128 \\
& Attention heads & 8 & 8 \\
& Encoder layers & 4 & 2 \\
& Feed-forward dimension & 512 & 512 \\
\midrule
Patch-based
& Patch length & 16 & 16 \\
& Patch stride & 8 & 8 \\
\midrule
All models
& Dropout & 0.15 & 0.15 \\
\bottomrule
\end{tabularx}
\end{table}

\exphead{KG-SSL comparison}
KG-SSL is evaluated only for RQ2 using its unsupervised
reconstruction-and-contrastive configuration without static-attribute
supervision~\citep{ghosh2022robust}. Its encoder is a one-layer bidirectional
LSTM with hidden dimension 64 and receives the same five meteorological
forcings together with observed streamflow. KG-SSL is trained using data from
1990--2009, after which one fixed descriptor is constructed for each basin and
used with the matched downstream LSTM at $L=365$. The descriptor remains
frozen during evaluation over 2010--2014. KG-SSL is not evaluated in RQ3
because constructing its representation requires observed streamflow from the
target basin, which is unavailable under zero-shot ungauged-basin transfer.

\exphead{Evaluation metrics}
Predictions are evaluated separately for each basin on the original,
unstandardized runoff scale. For basin $b$, with observed streamflow $q_{b,t}$
and predicted streamflow $\widehat{q}_{b,t}$ over $T_b$ evaluation days, the
Nash--Sutcliffe efficiency is
\begin{equation}
\operatorname{NSE}_b
=
1-
\frac{
\sum_{t=1}^{T_b}
\left(q_{b,t}-\widehat{q}_{b,t}\right)^2
}{
\sum_{t=1}^{T_b}
\left(q_{b,t}-\overline{q}_b\right)^2
},
\label{eq:app_nse}
\end{equation}
where $\overline{q}_b$ is the mean observed streamflow. The Kling--Gupta
efficiency is
\begin{equation}
\operatorname{KGE}_b
=
1-
\sqrt{
(r_b-1)^2+
(\alpha_b-1)^2+
(\beta_b-1)^2
},
\label{eq:app_kge}
\end{equation}
where
\begin{equation}
r_b
=
\operatorname{corr}(q_b,\widehat{q}_b),
\qquad
\alpha_b
=
\frac{\sigma_{\widehat{q}_b}}{\sigma_{q_b}},
\qquad
\beta_b
=
\frac{\overline{\widehat{q}}_b}{\overline{q}_b}.
\label{eq:app_kge_terms}
\end{equation}
Here, $r_b$ measures temporal correlation, $\alpha_b$ measures variability
bias, and $\beta_b$ measures mean-flow bias. Both metrics have an optimum of
$1$, with larger values indicating better performance
~\citep{Nash1970NSE,Gupta2009KGE}.

For RQ2, each seed is summarized by the median metric across the 671 evaluation
basins. For RQ3, predictions from all six held-out folds are combined so that
each basin contributes once, and the median is then computed across the full
held-out-basin population. In both settings, tables report the mean and
standard deviation of these medians over the three seeds. Fold-specific RQ3
results are reported separately in Appendix~A.4.

\exphead{Aggregate improvement summaries}
For each architecture $a$, let $s_a$ be the Basin2Vec score and $b_a$ the highest score among the alternative descriptors for the same metric and forcing window. The mean relative improvement is $100\,\frac{1}{5}\sum_{a=1}^{5}(s_a/b_a-1)$. The largest of the four temporal summaries (NSE/KGE, $L=120/365$) is $8.339\%$ for KGE with $L=365$ in Table~2, rounded to $8.3\%$. For spatial NSE in Table~3, the mean ratio $\frac{1}{5}\sum_{a=1}^{5}s_a/b_a$ is $3.5166$, rounded to $3.5$. These are arithmetic summaries of the reported basin-median scores across architectures, not ratios of prediction errors or measures of statistical significance.

\subsection{Extended Results for RQ1--RQ3}
\label{app:extended_results}

This section provides complementary analyses for the three research questions.
For RQ1, reference-basin similarity maps illustrate the spatial organization of
the Basin2Vec embedding space. For RQ2, basin-level distributions are reported
for both forcing-window lengths. For RQ3, we report fold-level results and
basin-level distributions for the representative $L=365$ setting. Because
higher NSE and KGE values indicate better performance, CDFs shifted farther to
the right represent stronger performance across a larger fraction of basins.

\subsubsection{Reference-Basin Embedding Similarity (RQ1)}
\label{app:rq1_reference_similarity}

To qualitatively examine the organization of the Basin2Vec embedding space, we
compute the cosine similarity between selected reference-basin descriptors and
all other basin descriptors. For reference basin $r$ and basin $b$, the
similarity is
\begin{equation}
s_{r,b}
=
\frac{
\bar{z}_r^{\top}\bar{z}_b
}{
\lVert \bar{z}_r \rVert_2
\lVert \bar{z}_b \rVert_2
},
\label{eq:app_reference_similarity}
\end{equation}
where $\bar{z}_r$ and $\bar{z}_b$ are the normalized basin-level descriptors
defined in Eq.14.

\begin{figure}[!htbp]
\centering
\includegraphics[width=\textwidth]
{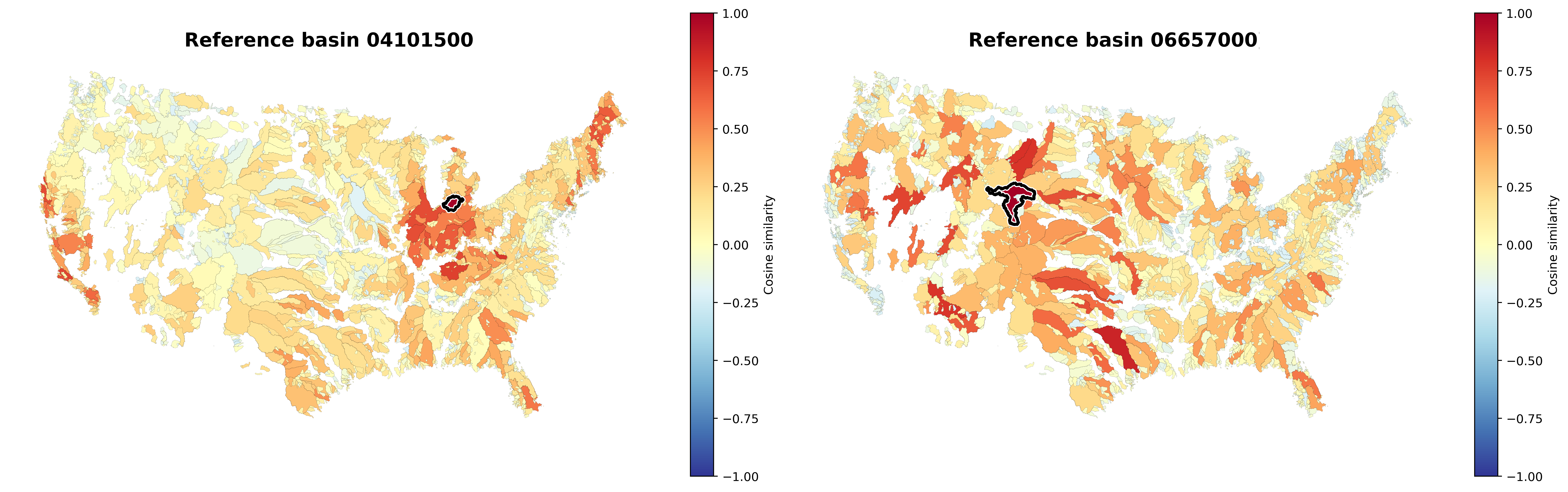}
\caption{
Example reference-basin similarity maps for RQ1. Each panel shows the cosine
similarity between one selected reference-basin descriptor and all Basin2Vec
basin descriptors. The reference basin is outlined in black. Similar basins
occur both near and far from each reference, indicating that the learned
neighborhoods are not determined solely by geographic proximity.
}
\label{fig:app_rq1_reference_similarity}
\end{figure}

Figure~\ref{fig:app_rq1_reference_similarity} shows distinct similarity
patterns for the two reference basins. Some geographically distant basins have
high similarity, whereas some nearby basins have lower similarity. These maps
qualitatively complement the effective-rank and metadata-consistency results in
the main paper by illustrating how individual basin neighborhoods vary across
the embedding space.

\subsubsection{Basin-Level Temporal Generalization (RQ2)}
\label{app:rq2_temporal_cdf}

\begin{figure}[!htbp]
\centering
\includegraphics[
    width=\textwidth,
    trim={0.25cm 0.2cm 0.1cm 0.2cm},
    clip
]{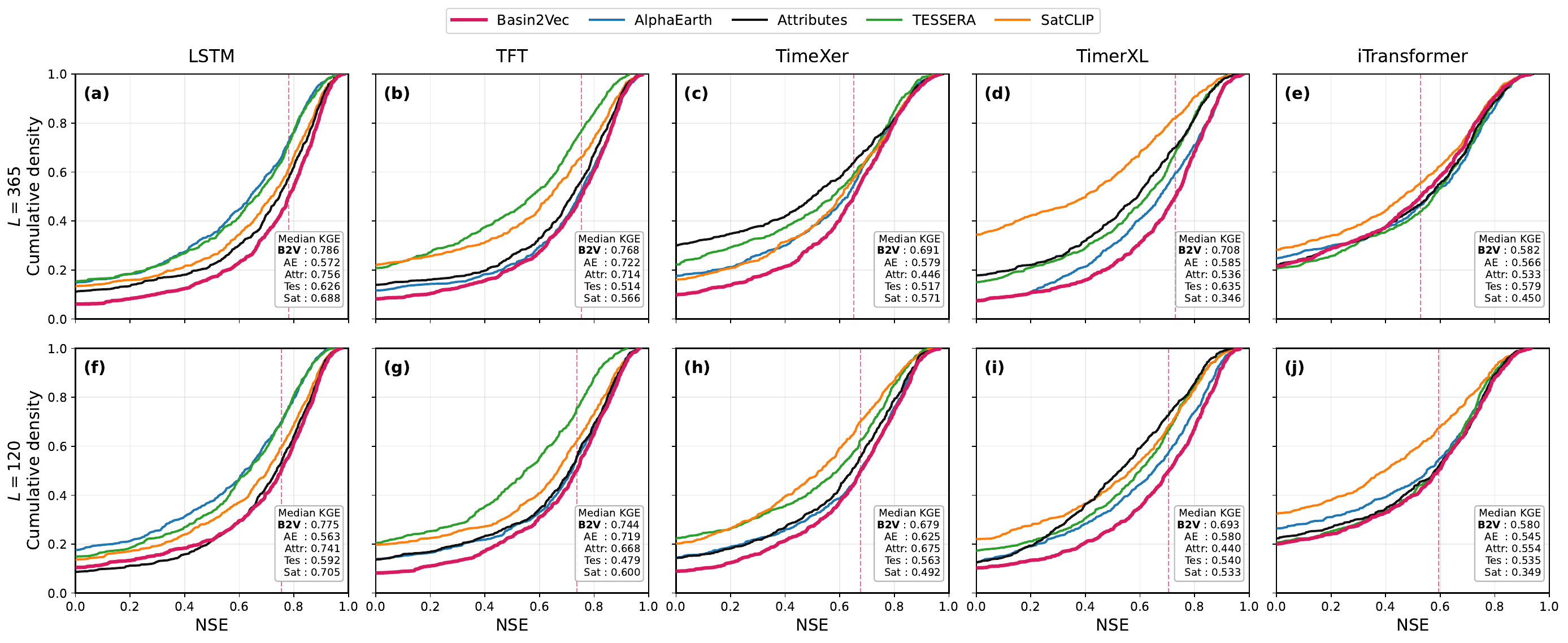}
\caption{
Basin-level performance distributions for RQ2 temporal generalization. Rows
correspond to forcing-window lengths $L=365$ and $L=120$, and columns
correspond to downstream forecasting architectures. The CDFs report NSE,
while the accompanying boxplots summarize KGE across the same evaluation
basins.
}
\label{fig:app_rq2_temporal_cdf}
\end{figure}

Figure~\ref{fig:app_rq2_temporal_cdf} complements the aggregate RQ2 results
with basin-level NSE distributions. The top and bottom rows correspond to
$L=365$ and $L=120$, respectively, and each column represents one downstream
forecasting architecture. The accompanying boxplots summarize basin-level KGE,
while the dashed vertical line in each CDF marks the Basin2Vec median.

The distributions show whether differences in median performance reflect
broad improvements across the basin population or gains concentrated in a
smaller subset of basins. The rightward shift of the Basin2Vec curves across
architectures and forcing-window lengths supports the aggregate temporal
generalization results reported in the main paper.

\subsubsection{Basin-Level Zero-Shot Spatial Transfer (RQ3)}
\label{app:rq3_spatial_cdf}

Table~\ref{tab:spatial_per_fold_compact} reports fold-level zero-shot
spatial-transfer results for $L=365$. Each entry gives the median NSE/KGE
across the held-out basins in one fold. Basin2Vec achieves the strongest
performance in every fold and downstream architecture, showing that its
spatial-transfer gains are not driven by a single basin partition.

\begin{table}[!htbp]
\centering
\caption{
Fold-level zero-shot spatial-transfer results for $L=365$. Each entry reports
median NSE/KGE across the unseen basins in that fold. B2V, AE, Attr., TES.,
and Sat. denote Basin2Vec, AlphaEarth, CAMELS attributes, TESSERA, and SatCLIP,
respectively. Bold indicates the best result.
}
\label{tab:spatial_per_fold_compact}

\begingroup
\scriptsize
\setlength{\tabcolsep}{0.35pt}
\renewcommand{\arraystretch}{1.10}

\resizebox{\columnwidth}{!}{%
\begin{adjustbox}{max width=\linewidth}
\begin{adjustbox}{max width=\linewidth}
\begin{tabular}{cl|ccccc}
\hline
Fold & Model & B2V & AE & Attr. & TES. & Sat. \\
\hline

\multirow{5}{*}{0}
& LSTM
& \textbf{0.479/0.381} & 0.179/$-0.064$ & 0.093/$-0.082$
& 0.081/$-0.127$ & 0.219/$-0.016$ \\
& TFT
& \textbf{0.516/0.420} & 0.209/0.004 & 0.253/0.024
& 0.127/$-0.090$ & 0.154/$-0.079$ \\
& TimeXer
& \textbf{0.489/0.408} & 0.088/$-0.052$ & 0.010/$-0.095$
& $-0.270$/$-0.319$ & 0.211/0.019 \\
& Timer-XL
& \textbf{0.515/0.504} & 0.181/$-0.038$ & 0.207/$-0.009$
& $-0.146$/$-0.123$ & 0.112/$-0.022$ \\
& iTrans.
& \textbf{0.440/0.411} & 0.200/0.010 & 0.056/$-0.011$
& 0.077/$-0.166$ & 0.123/$-0.065$ \\
\hline

\multirow{5}{*}{1}
& LSTM
& \textbf{0.434/0.442} & 0.162/$-0.008$ & 0.136/$-0.028$
& $-0.004$/$-0.118$ & 0.017/$-0.047$ \\
& TFT
& \textbf{0.495/0.365} & 0.225/0.038 & 0.110/$-0.126$
& 0.204/0.016 & 0.206/$-0.062$ \\
& TimeXer
& \textbf{0.466/0.452} & 0.240/0.126 & 0.095/$-0.041$
& 0.187/0.000 & 0.140/$-0.032$ \\
& Timer-XL
& \textbf{0.486/0.402} & 0.207/0.088 & 0.133/0.016
& 0.020/$-0.081$ & 0.171/0.048 \\
& iTrans.
& \textbf{0.421/0.362} & 0.239/0.103 & 0.043/$-0.096$
& 0.157/$-0.045$ & 0.017/$-0.142$ \\
\hline

\multirow{5}{*}{2}
& LSTM
& \textbf{0.437/0.333} & $-0.029$/$-0.147$ & 0.000/$-0.165$
& 0.084/$-0.057$ & 0.050/$-0.098$ \\
& TFT
& \textbf{0.499/0.374} & $-0.041$/$-0.146$ & 0.158/$-0.118$
& 0.190/$-0.034$ & 0.065/$-0.144$ \\
& TimeXer
& \textbf{0.480/0.386} & 0.050/$-0.143$ & 0.061/$-0.070$
& 0.026/$-0.097$ & 0.000/$-0.136$ \\
& Timer-XL
& \textbf{0.482/0.383} & 0.070/$-0.066$ & $-0.040$/$-0.122$
& 0.025/$-0.182$ & $-0.036$/$-0.141$ \\
& iTrans.
& \textbf{0.479/0.339} & $-0.083$/$-0.166$ & $-0.003$/$-0.141$
& 0.071/$-0.166$ & $-0.077$/$-0.218$ \\
\hline

\multirow{5}{*}{3}
& LSTM
& \textbf{0.466/0.415} & 0.089/$-0.055$ & 0.082/$-0.047$
& 0.105/$-0.001$ & 0.015/$-0.125$ \\
& TFT
& \textbf{0.515/0.425} & 0.136/$-0.041$ & 0.113/$-0.084$
& 0.102/$-0.022$ & 0.114/$-0.025$ \\
& TimeXer
& \textbf{0.494/0.450} & 0.205/0.083 & 0.054/$-0.029$
& 0.092/0.051 & 0.062/$-0.045$ \\
& Timer-XL
& \textbf{0.480/0.412} & 0.198/0.045 & 0.069/$-0.003$
& 0.051/$-0.088$ & 0.160/0.016 \\
& iTrans.
& \textbf{0.436/0.343} & 0.171/0.079 & $-0.099$/$-0.104$
& 0.105/$-0.063$ & 0.075/$-0.036$ \\
\hline

\multirow{5}{*}{4}
& LSTM
& \textbf{0.438/0.407} & 0.001/$-0.098$ & $-0.029$/$-0.177$
& 0.085/$-0.021$ & 0.215/$-0.028$ \\
& TFT
& \textbf{0.533/0.489} & 0.121/$-0.136$ & 0.047/$-0.140$
& 0.096/$-0.101$ & 0.046/$-0.162$ \\
& TimeXer
& \textbf{0.496/0.427} & 0.043/$-0.065$ & $-0.121$/$-0.150$
& 0.071/$-0.063$ & 0.141/$-0.058$ \\
& Timer-XL
& \textbf{0.477/0.443} & 0.142/0.005 & $-0.082$/$-0.192$
& $-0.048$/$-0.154$ & 0.147/$-0.094$ \\
& iTrans.
& \textbf{0.481/0.390} & 0.130/$-0.107$ & $-0.162$/$-0.113$
& 0.054/$-0.097$ & 0.088/$-0.119$ \\
\hline

\multirow{5}{*}{5}
& LSTM
& \textbf{0.496/0.443} & $-0.006$/$-0.155$ & 0.086/$-0.090$
& 0.193/0.014 & 0.110/$-0.135$ \\
& TFT
& \textbf{0.564/0.515} & 0.113/$-0.163$ & 0.198/$-0.024$
& 0.165/$-0.088$ & 0.165/$-0.074$ \\
& TimeXer
& \textbf{0.527/0.521} & 0.041/$-0.116$ & 0.129/$-0.097$
& 0.120/0.004 & 0.028/$-0.109$ \\
& Timer-XL
& \textbf{0.542/0.457} & 0.028/$-0.208$ & 0.160/$-0.130$
& 0.025/$-0.068$ & 0.175/$-0.024$ \\
& iTrans.
& \textbf{0.519/0.433} & 0.138/$-0.077$ & 0.107/$-0.059$
& 0.097/$-0.086$ & 0.025/$-0.129$ \\
\hline
\end{tabular}
\end{adjustbox}
\end{adjustbox}%
}

\endgroup
\end{table}

\begin{figure}[!htbp]
\centering
\includegraphics[
    width=\textwidth,
    trim={0.25cm 0.2cm 0.1cm 0.2cm},
    clip
]{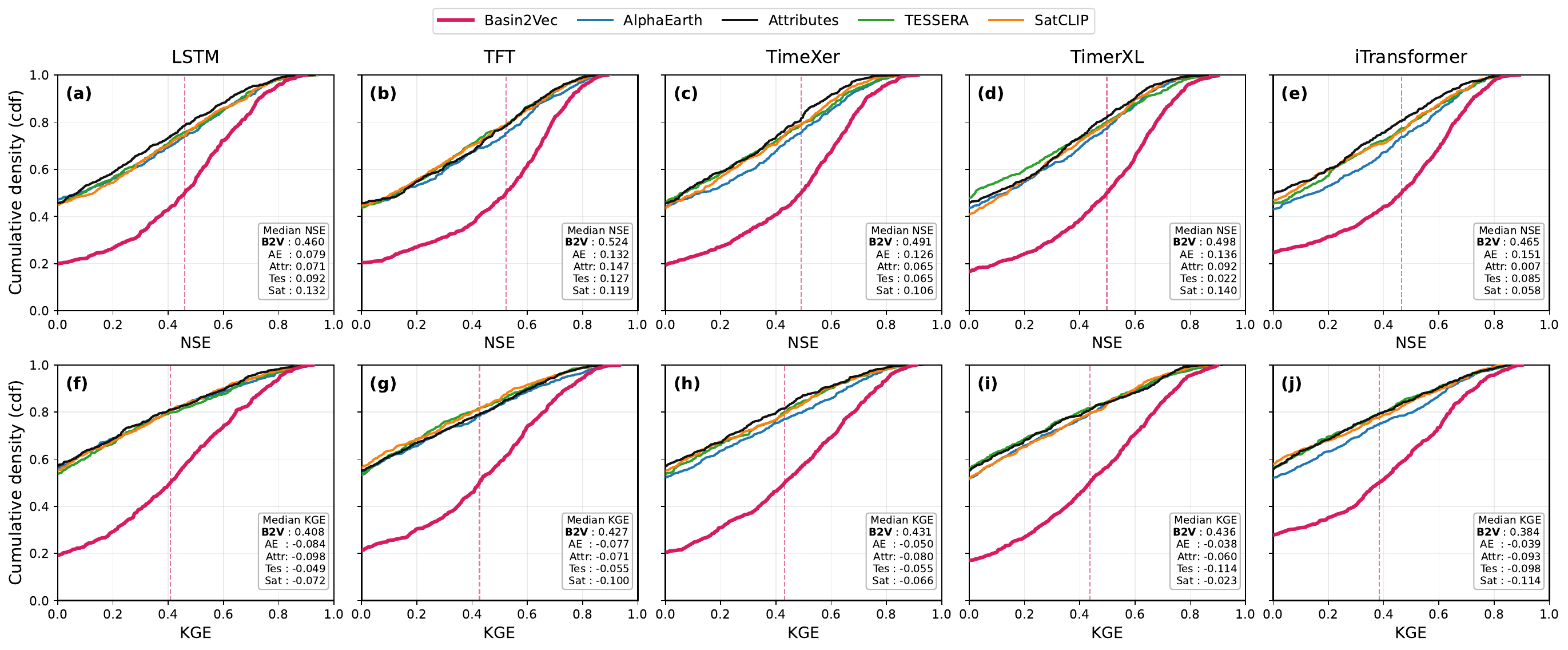}
\caption{
Basin-level performance distributions for RQ3 zero-shot spatial transfer at
$L=365$. The top row reports NSE and the bottom row reports KGE across held-out
basins. Columns correspond to downstream forecasting architectures, and curves
compare Basin2Vec, AlphaEarth, CAMELS attributes, TESSERA, and SatCLIP.
}
\label{fig:app_rq3_spatial_cdf}
\end{figure}

Figure~\ref{fig:app_rq3_spatial_cdf} complements the fold-level results with
basin-level NSE and KGE distributions aggregated across the held-out basins
for $L=365$. The top row reports NSE, the bottom row reports KGE, and each
column corresponds to a downstream forecasting architecture. The dashed
vertical line marks the Basin2Vec median in each panel.

Together, the fold-level results and CDFs show that Basin2Vec's improvements
persist across alternative basin partitions and throughout the held-out-basin
population, including the lower-performing tail.

\paragraph{Why does the LSTM outperform the attention-based predictors?}
The LSTM generally performs better than the attention-based architectures in
our experiments. This is consistent with large-sample hydrology studies showing
that LSTMs provide a strong inductive bias for rainfall--runoff
modeling~\citep{Kratzert2019EntityAwareLSTM}. Attention-based models can capture
long-range dependencies, but their advantage is not universal and depends on
the task and data regime~\citep{Lim2021TFT,Zeng2023Transformers}. In our
setting, the forcing history is limited to $L\in\{120,365\}$ days, and
basin-specific information is already supplied through a fixed descriptor.
This reduces the need for the downstream predictor to infer long-range basin
context solely from meteorological inputs. We therefore interpret the LSTM's
stronger performance as specific to this experimental setting rather than a
general superiority over attention-based architectures.

\clearpage
\subsection{Metadata Conditioning and Grouped Objective Ablations}
\label{app:factorial}

\exphead{Configuration identity}
Table~4 provides the authoritative values for the configurations summarized here. Its ``hard positive only'' variant retains direct metadata conditioning while using augmentation positives and uniform contrastive-denominator weights. Accordingly, this row is labeled Base + metadata in Table~\ref{tab:factorial_results}, rather than Base. Its temporal NSE/KGE is $0.670/0.647$ and spatial NSE/KGE is $0.372/0.332$. The separately supplied $0.68/0.69$ temporal and $0.43/0.38$ spatial summaries are not used for this configuration because their correspondence to a distinct run configuration has not been established. No verified result is available for Base without direct metadata conditioning; its cells remain unreported.

\exphead{Scope of the grouped comparison}
Table~\ref{tab:factorial_switches} distinguishes the reported positive-pair and denominator rules. Other objective terms follow Table~4. In particular, the ``hard positive only'' row alone does not establish that the negative-margin objective was disabled. These rows therefore do not constitute a completed, strictly controlled two-by-two factorial experiment. Base + relations corresponds to Table~4's variant without metadata conditioning, while Full uses all components. Area regularization remains active in these Table~4 variants. Polygon-based overlap relations are distinct from the constant-zero overlap-summary input fields in Table~\ref{tab:basin2vec_metadata}.

\exphead{Evaluation and interpretation}
The reported Table~4 results use LSTM with $L=365$ under temporal and six-fold spatial evaluation, summarized as mean $\pm$ standard deviation across three seeds. Full exceeds the metadata-conditioned hard-positive control by $0.112$ NSE and $0.139$ KGE temporally, and $0.088$ NSE and $0.076$ KGE spatially. Relative to the variant without direct metadata conditioning, the gains are $0.148/0.171$ temporally and $0.396/0.467$ spatially. These observations support the importance of metadata conditioning and the benefit of the complete objective over the hard-positive control. They do not isolate an individual loss component or establish statistical significance.

\exphead{Limits}
Table~\ref{tab:factorial_effects} gives arithmetic differences between the central values in Table~4. No factorial interaction is estimated because the Base cell is unreported, and paired uncertainty for the differences cannot be recovered from marginal standard deviations alone. Every reported variant retains area supervision, and the relational objective can use metadata; ``without direct metadata conditioning'' is not equivalent to metadata-free learning. This comparison also does not establish metadata parity for external EO or CAMELS descriptors.

\begin{table}[!htbp]
\centering\small
\caption{Configuration identities for the grouped comparison. Reported configurations and objective terms follow Table~4. Base is unreported and is not assigned an inferred result.}
\label{tab:factorial_switches}
\begin{adjustbox}{max width=\linewidth}
\begin{tabular*}{\textwidth}{@{\extracolsep{\fill}}lccc@{}}
\toprule Configuration & Direct metadata & Positive-pair rule & Denominator weights\\\midrule
Base & Off & Not evaluated here & Not evaluated here\\
Base + metadata & On & Hard positives only & Uniform\\
Base + relations & Off & Full relation-aware positives & Metadata-weighted\\
Full Basin2Vec & On & Full relation-aware positives & Metadata-weighted\\\bottomrule
\end{tabular*}
\end{adjustbox}
\end{table}

\begin{table}[!htbp]
\centering\small
\caption{Grouped ablation results from Table~4 using LSTM with $L=365$. Values are mean $\pm$ standard deviation across three seeds; spatial evaluation uses six folds. Base + metadata is Table~4's ``hard positive only'' variant; Base + relations is its ``without metadata conditioning'' variant. A dash indicates an unreported result, not zero.}
\label{tab:factorial_results}
\begin{adjustbox}{max width=\linewidth}
\begin{tabular*}{\textwidth}{@{\extracolsep{\fill}}lcccc@{}}
\toprule & \multicolumn{2}{c}{Temporal} & \multicolumn{2}{c}{Spatial}\\
\cmidrule(lr){2-3}\cmidrule(lr){4-5}
Configuration & NSE $\uparrow$ & KGE $\uparrow$ & NSE $\uparrow$ & KGE $\uparrow$\\\midrule
Base & -- & -- & -- & --\\
Base + metadata & $0.670\pm0.009$ & $0.647\pm0.021$ & $0.372\pm0.010$ & $0.332\pm0.021$\\
Base + relations & $0.634\pm0.022$ & $0.615\pm0.017$ & $0.064\pm0.044$ & $-0.059\pm0.044$\\
Full Basin2Vec & $0.782\pm0.009$ & $0.786\pm0.011$ & $0.460\pm0.050$ & $0.408\pm0.001$\\\bottomrule
\end{tabular*}
\end{adjustbox}
\end{table}

\begin{table}[!htbp]
\centering\small
\caption{Descriptive Full-model gains calculated from the central values in Table~4. No significance or factorial interaction is inferred.}
\label{tab:factorial_effects}
\begin{adjustbox}{max width=\linewidth}
\begin{tabular*}{\textwidth}{@{\extracolsep{\fill}}lcccc@{}}
\toprule & \multicolumn{2}{c}{Temporal} & \multicolumn{2}{c}{Spatial}\\
\cmidrule(lr){2-3}\cmidrule(lr){4-5}
Contrast & $\Delta$NSE & $\Delta$KGE & $\Delta$NSE & $\Delta$KGE\\\midrule
Full $-$ metadata-conditioned hard-positive control & +0.112 & +0.139 & +0.088 & +0.076\\
Full $-$ without-metadata-conditioning control & +0.148 & +0.171 & +0.396 & +0.467\\\bottomrule
\end{tabular*}
\end{adjustbox}
\end{table}
\clearpage
% ------------------------ References ------------------------

\end{document}